%% file: root.tex
\documentclass[letterpaper, 10 pt, journal, twoside]{IEEEtran}

\IEEEoverridecommandlockouts

\input{macros.tex}

\title{
HEIGHT: Heterogeneous Interaction 
Graph Transformer for \\ Robot Navigation in Crowded and Constrained Environments
}

\author{

Shuijing Liu\textsuperscript{1}, 
Haochen Xia\textsuperscript{2,*}, 
Fatemeh Cheraghi Pouria\textsuperscript{2,*}, 
Kaiwen Hong\textsuperscript{2}, \\
Neeloy Chakraborty\textsuperscript{2},
Zichao Hu\textsuperscript{1}, 
Joydeep Biswas\textsuperscript{1}, 
Katherine Driggs-Campbell\textsuperscript{2} \\
\vspace{5pt}
\textsuperscript{1}The University of Texas at Austin, 
\textsuperscript{2}University of Illinois Urbana-Champaign. \\
\textsuperscript{*}Equal contribution
}

\begin{document}
\twocolumn[{%
	\renewcommand\twocolumn[1][]{#1}%
	\maketitle
        \vspace{-6mm}

    \centering
      \includegraphics[width=1.0\linewidth]{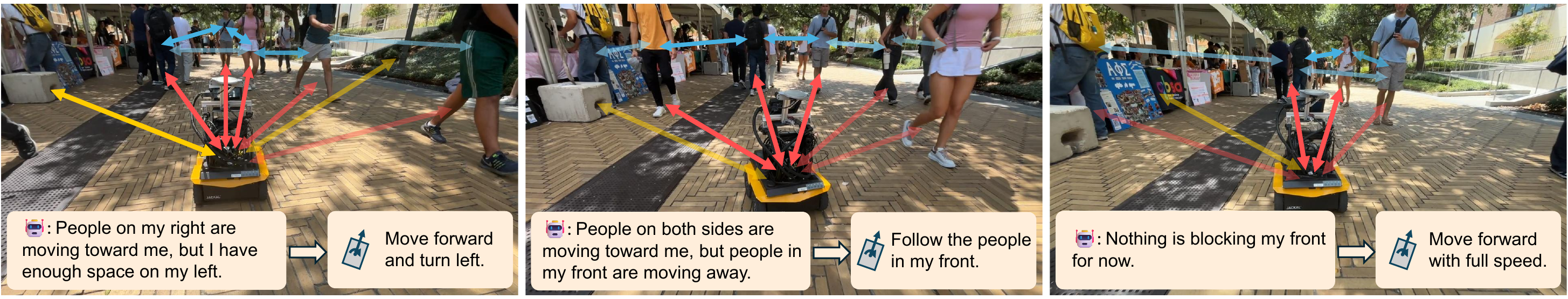}
		\captionof{figure}{\textbf{A heterogeneous graph aids the robot's spatio-temporal reasoning in a crowded and constrained environment.} The colored arrows denote \textcolor{red}{robot-human (RH)}, \textcolor{blue}{human-human (HH)}, and \textcolor{YellowOrange}{obstacle-agent (OA)} interactions. The opaque arrows are the more important interactions, while the transparent arrows are the less important ones. The white arrows indicate the front direction of the robot and its length indicates the robot speed. At each timestep $t$, the robot reasons about these interactions, focuses on the important ones, and makes decisions.
        } \label{fig:open}
    \vspace{2mm}
}]

\begingroup
\replaythanks{Manuscript received July, 10, 2023; Revised August 8, 2025 and October 12, 2025; Accepted November 13, 2025; Date of publication December 15, 2025; Date of current version December 15, 2025. This article was recommended for publication by
Associate Editor and Editor Chenguang Yang upon evaluation of
the reviewers’ comments.}
\replaythanks{Corresponding author: Shuijing Liu (shuijing.liu@utexas.edu).}

\replaythanks{Digital Object Identifier (DOI): 10.1109/TASE.2025.3646588.}

\endgroup

\begin{abstract}

We study the problem of robot navigation in dense and interactive crowds with static constraints such as corridors and furniture. 
Previous methods fail to consider all types of spatial and temporal interactions among agents and obstacles, leading to unsafe and inefficient robot paths.
In this article, we leverage a graph-based representation of crowded and constrained scenarios and propose a structured framework to learn robot navigation policies with deep reinforcement learning.
We first split the representations of different inputs, and propose a
heterogeneous spatio-temporal (st) graph to model distinct interactions among humans, robots, and obstacles.
Based on the \hstGraph, we propose \modelName, a novel navigation policy network architecture with different components to capture heterogeneous interactions through space and time. 
HEIGHT utilizes attention mechanisms to prioritize important interactions and a recurrent network to track changes in the dynamic scene over time, encouraging the robot to avoid collisions adaptively.
Through extensive simulation and real-world experiments, we demonstrate that \modelName outperforms state-of-the-art baselines in terms of success, navigation time, and generalization to domain shifts in challenging navigation scenarios.
More information is available at {\color{cyan}{\url{https://sites.google.com/view/crowdnav-height/home}}}.
\end{abstract}

\makeatletter
\def\notetopractitioners{\normalfont%
  \if@twocolumn
    \@IEEEabskeysecsize\bfseries\textit{Note to Practitioners}---\,%
  \else
    \begin{center}\vspace{-1.78ex}\@IEEEabskeysecsize\textbf{Note to Practitioners}\end{center}\quotation\@IEEEabskeysecsize%
  \fi\@IEEEgobbleleadPARNLSP}
\def\endnotetopractitioners{\relax\vspace{1.34ex}\par\if@twocolumn\else\endquotation\fi%
  \normalfont\normalsize}
\makeatother


\begin{notetopractitioners}
This work is motivated by the challenge of safe and efficient robot navigation in indoor environments with people, such as homes and offices. Conventional navigation methods make simplistic assumptions about interactions among agents and entities, which leads to unsafe or unnatural robot behavior in dynamic settings. Previous machine learning methods either ignore or use ineffective approaches to handle static obstacles—limiting their real-world applicability. For a practical system to reason about both dynamic people and static constraints simultaneously, the following two factors are critical: how the robot represents its surroundings and how it make decisions. To represent the dynamic environment, our method treats people and obstacles differently and keeps the most essential information as inputs. On the architectural side, we propose a structured graph-based network, which consists of specialized models to explicitly models different types of interactions among all observed entities.
In challenging simulation benchmarks, our approach outperforms alternative designs that are commonly used in existing works. We also demonstrate strong real-world deployment in everyday environments. 
To improve deployment, a future direction is continual learning in the real-world.
\end{notetopractitioners}


\input{Sections/01-Introduction.tex}

\input{Sections/02-LitReview.tex}

\input{Sections/03-Preliminaries}

\input{Sections/04-Methods.tex}

\input{Sections/05-SimExp.tex}

\input{Sections/06-RealExp.tex}

\input{Sections/07-Discussion.tex}

\input{Sections/08-Conclusion.tex}



\bibliographystyle{IEEEtran}
\bibliography{BibFile}
\input{Sections/Appendix}

\newpage
\begin{IEEEbiography}[{\includegraphics[width=1in,height=1.25in,clip,keepaspectratio]{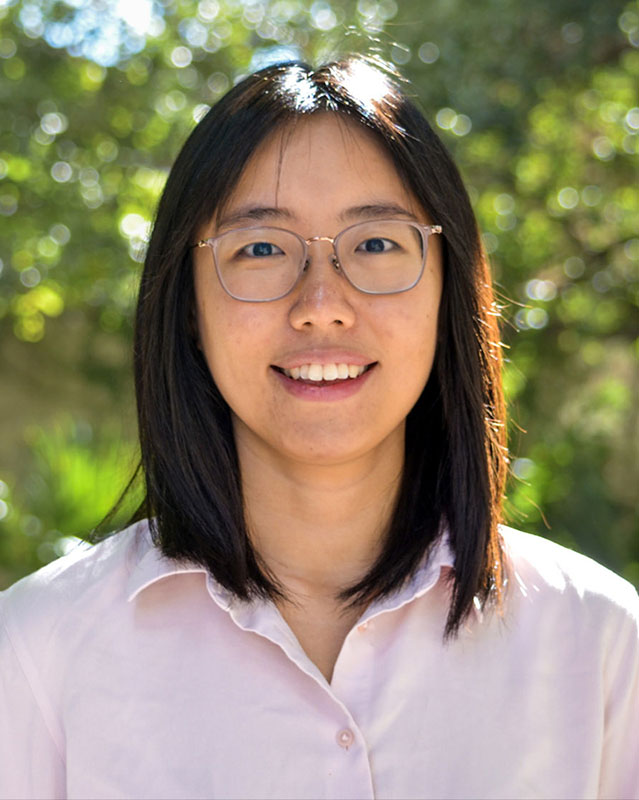}}]{Shuijing Liu}
received B.S. and Ph.D. degrees in Electrical and Computer Engineering from University of Illinois Urbana-Champaign in 2018 and 2024, respectively. She is currently a postdoctoral researcher in the Computer Science Department at The University of Texas at Austin. She also serves on the Artificial Intelligence advisory board of a startup named Municorn.ai. Her research interest is at the intersection of human-robot interaction and robot learning, with the goal of building a trustworthy partnership between humans and robots in the wild. 
\end{IEEEbiography}

\begin{IEEEbiography}
[{\includegraphics[width=1in,height=1.25in,clip,keepaspectratio]{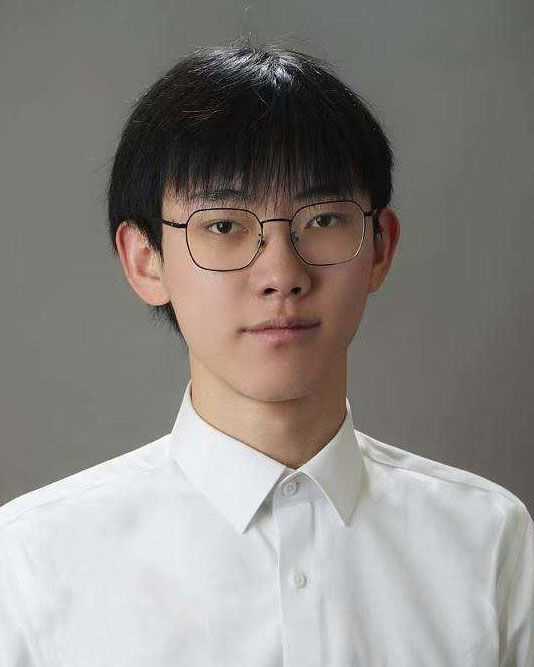}}]{Haochen Xia}
is currently pursuing a B.S. degree in computer engineering, with a minor in mathematics, at the University of Illinois at Urbana–Champaign, Urbana, IL, USA. He is an undergraduate Research Assistant with the Human-Centered Autonomy Lab and the ECE Department, working under Prof. Katie Driggs-Campbell and Prof. David M. Nicol. His research interests include human–robot interaction, cognitive AI, multimodal perception, and learning-based motion planning.
\end{IEEEbiography}

\begin{IEEEbiography}
[{\includegraphics[width=1in,height=1.25in,clip,keepaspectratio]{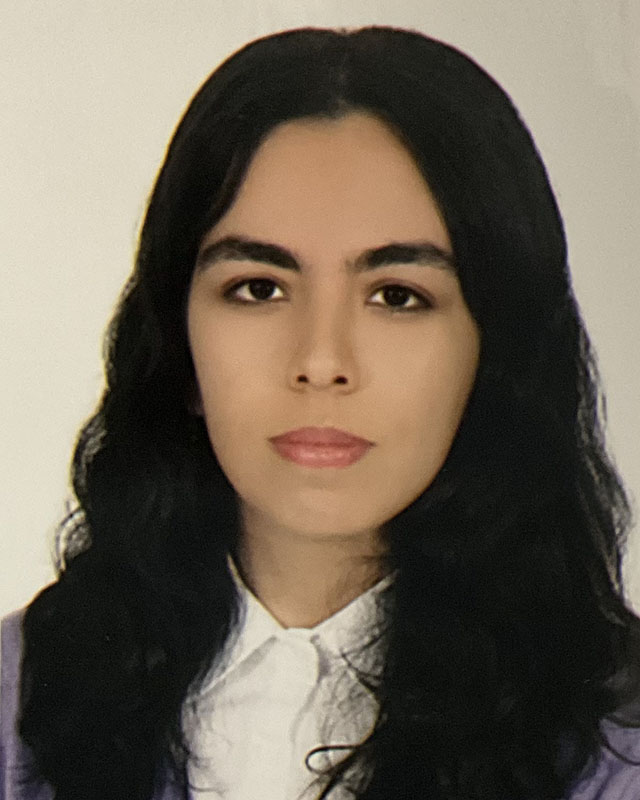}}]{Fatemeh Cheraghi Pouria}
received a B.S. degree in Mechanical Engineering from Sharif University of Technology, Tehran, Iran, in 2022. She is currently pursuing a Ph.D. degree in Electrical and Computer Engineering at University of Illinois Urbana-Champaign.
Her current research interests include Human Trajectory Prediction and Robot navigation.
\end{IEEEbiography}

\begin{IEEEbiography}
[{\includegraphics[width=1in,height=1.25in,clip,keepaspectratio]{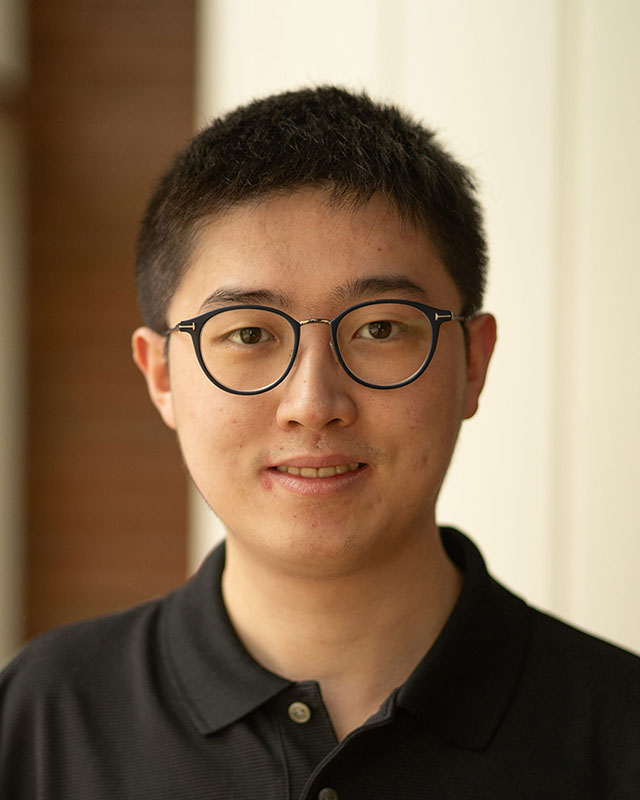}}]{Kaiwen Hong}
received a B.E. degree in electrical engineering from Zhejiang University, Hangzhou, China, and the B.S. degree in electrical engineering from University of Illinois at Urbana-Champaign, Urbana, IL, USA, in 2021. He is currently pursuing a Ph.D. degree in electrical and computer engineering at the University of Illinois at Urbana-Champaign.
His current research interests include robot learning and human-centered robotics.
\end{IEEEbiography}

\begin{IEEEbiography}
[{\includegraphics[width=1in,height=1.25in,clip,keepaspectratio]{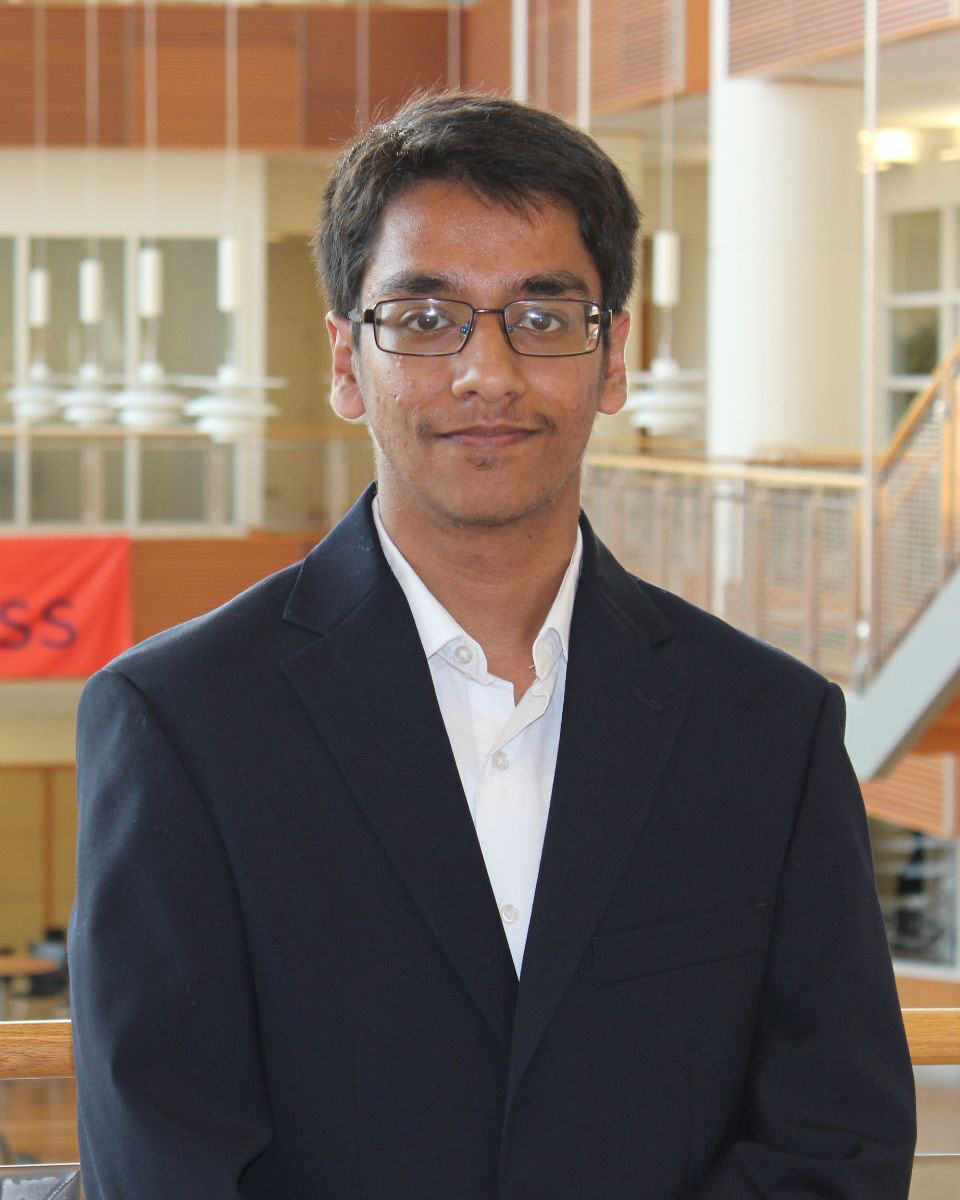}}]{Neeloy Chakraborty}
received B.S. and M.S. degrees in Computer Engineering from University of Illinois at Urbana-Champaign in 2021 and 2013, respectively. He is currently pursuing a Ph.D. degree in Electrical and Computer Engineering at University of Illinois at Urbana-Champaign.
His current research interests include foundation model evaluation and field robotics.
\end{IEEEbiography}

\begin{IEEEbiography}
[{\includegraphics[width=1in,height=1.25in,clip,keepaspectratio]{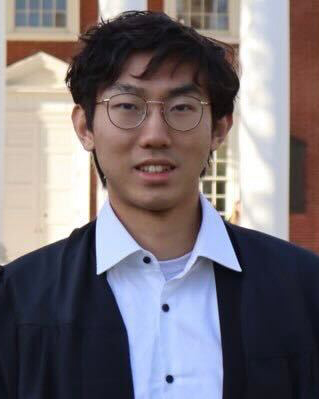}}]{Zichao Hu}
(Member, IEEE) received a B.S. degree in computer engineering from the University of Virginia, Virginia, USA, in 2022. He is currently pursuing a Ph.D. degree in computer science at the University of Texas at Austin, Texas, USA. His current research interests include diffusion policy for social navigation and whole-body control for humanoid robot object interaction.
\end{IEEEbiography}

\begin{IEEEbiography}
[{\includegraphics[width=1in,height=1.25in,clip,keepaspectratio]{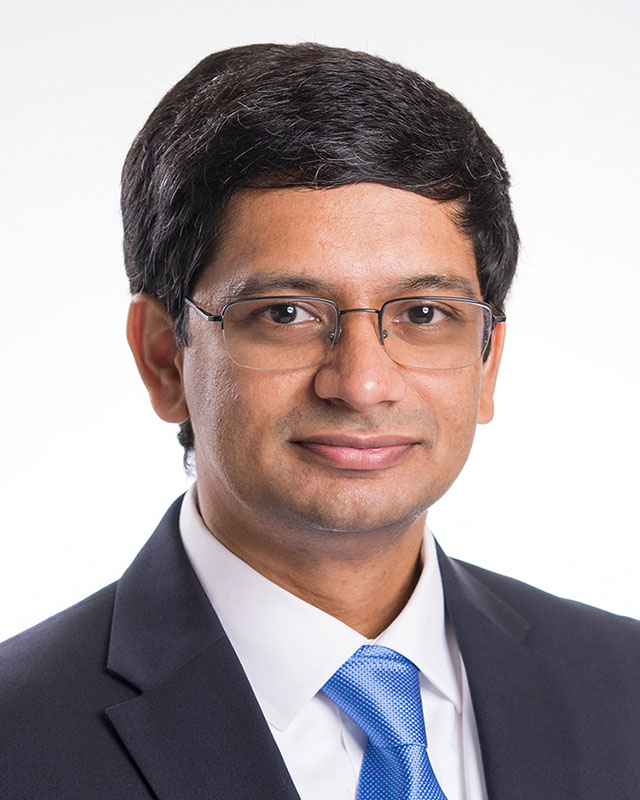}}]{Joydeep Biswas}
(Member, IEEE) received a B.Tech. degree in Engineering Physics from Indian Institute of Technology, Bombay in 2008, and the M.S. and Ph.D. degrees in Robotics from Carnegie Mellon University in 2010 and 2014, respectively. He is currently an Associate Professor in the Computer Science department at the University of Texas at Austin and a visiting professor at Nvidia. His research interests include robot perception, motion planning, control systems, AI,  and deployed robot systems.
\end{IEEEbiography}

\begin{IEEEbiography}
[{\includegraphics[width=1in,height=1.25in,clip,keepaspectratio]{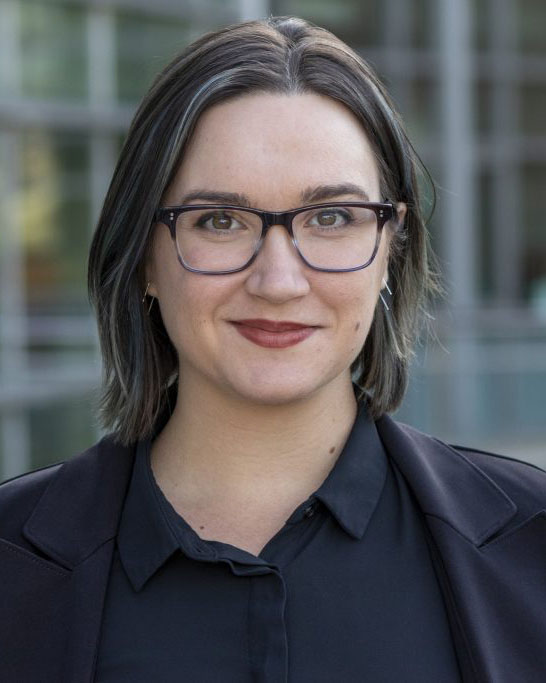}}]{Katherine Driggs-Campbell}
(Member, IEEE) received a B.S.E degree in Electrical Engineering from Arizona State University in 2012, and the M.S. and Ph.D. degrees in Electrical Engineering and Computer Sciences from University of California, Berkeley in 2015 and 2017, respectively. She is currently an Associate Professor in the Electrical and Computer Engineering department at University of Illinois at Urbana-Champaign. Her research interest is developing tools for autonomous robotics that operate out in the real world with people, which is drawn from the fields of human-robot interaction, optimization, learning \& AI, and control theory.
\end{IEEEbiography}
\end{document}

%% file: macros.tex
\usepackage[utf8]{inputenc}
\usepackage{amsmath,amssymb,stmaryrd,mathtools}
\usepackage{amsfonts}

\usepackage[noend]{algpseudocode}
\usepackage{acronym}
\usepackage{color}
\usepackage[dvipsnames]{xcolor}

\usepackage{verbatim}
\usepackage{booktabs}
\usepackage{siunitx}
\usepackage{graphics}
\usepackage{graphicx,caption}
\usepackage{suffix}
\usepackage{xstring}
\usepackage{xparse}
\usepackage{expl3}
\usepackage{mathrsfs}
\usepackage{tabularx}
\usepackage{makecell}
\usepackage{array}
\usepackage{hyperref}
\usepackage{cleveref}
\usepackage[ruled,linesnumbered]{algorithm2e}
\usepackage{multirow}
\usepackage{diagbox}
\usepackage{rotating}
\usepackage{bbm}
\usepackage{dsfont}

\usepackage{subcaption}
\usepackage{wrapfig}

\usepackage{tikz}
\usetikzlibrary{arrows,backgrounds,calc}
\usepackage{relsize}
\usepackage{float}
\usepackage{kantlipsum} 
\usepackage{lipsum}
\usepackage{stfloats}
\usepackage{siunitx}

\usepackage[noadjust]{cite}
\usepackage{todonotes}
\usepackage{soul}
\definecolor{smoothgreen}{rgb}{0.7,1,0.7}
\sethlcolor{smoothgreen}

\RequirePackage{luatex85}
\usepackage{pgfplots}
\pgfplotsset{compat=newest}
\pgfplotsset{every axis legend/.append style={%
		cells={anchor=west}}
}
\usepgfplotslibrary{polar}
\usetikzlibrary{arrows}
\tikzset{>=stealth'}

\definecolor{C1}{rgb}{0.0, 0.447, 0.741}
\definecolor{C1_light}{rgb}{0.0, 0.6032388663967612, 1.0}
\definecolor{C2}{rgb}{0.85, 0.325, 0.098}
\definecolor{C3}{rgb}{0.929, 0.694, 0.125}
\definecolor{C4}{rgb}{0.494, 0.184, 0.556}
\definecolor{C5}{rgb}{0.466, 0.674, 0.188}
\definecolor{C6}{rgb}{0.301, 0.745, 0.933}
\definecolor{C7}{rgb}{0.635, 0.078, 0.184}

\usepgfplotslibrary{groupplots}

\usetikzlibrary{shapes.geometric, arrows}

\tikzstyle{startstop} = [rectangle, rounded corners, minimum width=2cm, minimum height=1cm,text centered, draw=black, fill=none]
\tikzstyle{arrow} = [thick,->,>=stealth]

\newcommand{\modelNameUpper}{HEIGHT\xspace}
\newcommand{\modelName}{HEIGHT\xspace}
\newcommand{\astarCNN}{$\textrm{A}^{*}$+\,CNN\xspace}
\newcommand{\hstGraph}{heterogeneous st-graph\xspace}
\newcommand{\homoBaseline}{HomoGAT\xspace}

\newcommand{\replaythanks}[1]{%
  \begingroup
  \renewcommand\thefootnote{}%
  \footnotetext{#1}%
  \endgroup
}

%% file: Sections/01-Introduction.tex
\section{Introduction}
\label{sec:intro}
\IEEEPARstart{R}{obots} are increasingly prevalent in human-centric environments.
In addition to dynamic agents, real-world navigation environments usually contain static constraints such as walls, furniture, and untraversable terrains.
In this article, we study robot navigation to a destination without colliding with humans \textbf{and} obstacles, a crucial ability for applications such as last-mile delivery and household robots.
For example, Fig.~\ref{fig:open} shows such a navigation scenario with abundant subtle interactions among the robot, humans, and obstacles.
These interactions are heterogeneous, dynamic, and difficult to reason, making navigation in such environments challenging.



While a large body of research on robot crowd navigation studies open-world navigation and ignores static obstacles~\cite{chen2019crowd,liu2023intention,wang2023navistar}, 
some previous works have explored various approaches for crowd navigation in constrained spaces~\cite{fox1997dynamic,van2008reciprocal,perez2021robot}. 
However, these works typically have one of two limitations: (1) They do not differentiate between dynamic and static obstacles, and thus the robot has difficulties taking adaptive strategies to avoid collisions~\cite{fox1997dynamic,chen2019relational,perez2021robot,liu2020decentralized};
(2) The representations of scene, especially static obstacles, are inefficient or redundant as inputs to the navigation algorithm.
To address these limitations, we ask the following research question: \textit{How can a robot navigation policy represent and reason about diverse interactions in crowded and constrained environments to adaptively avoid collisions?}

To answer this question, we propose a framework that leverages the heterogeneity of interactions in crowded and constrained scenarios in both scene representation and navigation policy design. First, we use light weight perception modules to extract separate human and obstacle representations that are essential for navigation, such as relative positions w.r.t. the robot. The split representations are processed and fed separately into the reinforcement learning (RL)-based navigation pipeline.
Then, inspired by recent breakthroughs in spatio-temporal (st) networks for crowd navigation~\cite{liu2020decentralized,wang2023navistar,liu2023intention,chen2024socially}, we decompose the scenario into a heterogeneous st-graph with different types of edges to represent different types of interactions among the robot, detected humans, and observed obstacles, as shown in the colored arrows in Fig.~\ref{fig:open}.
Finally, we convert the \hstGraph into a \textbf{HE}terogeneous \textbf{I}nteraction \textbf{G}rap\textbf{H} \textbf{T}ransformer (HEIGHT), a robot policy network consisting of different modules to parameterize the various spatio-temporal interactions. 
Specifically, we use two separate attention networks to address the different effects of robot-human (RH) and human-human (HH) interactions. The attention networks enable the robot
to pay more attention to the important interactions, leading to a low collision-rate even as the number of humans increases.

In the rapidly changing scenario in Fig.~\ref{fig:open}, \modelName injects structures to and captures the synergy between scene representation and network architectures. By reasoning about the heterogeneous
interactions among different components through space and time, the robot is able to avoid collisions and approach its goal in an efficient manner. In summary, the main contributions of this article are as follows.


\begin{enumerate}
    \item We propose an input representation of crowded and constrained environments that treats observed humans and obstacles differently. The split scene representation naturally allows us to inject structures in the framework. 
    \item We propose a heterogeneous spatio-temporal (st) graph to effectively model the pairwise interactions among all agents and entities in crowded and constrained scenarios.
    \item From the heterogenous st-graph, we use a principle approach to derive \modelName, a transformer-based robot navigation policy network with different modules to reason about all types of spatial and temporal interactions. 
    \item In simulation benchmarks with dense crowds and dense obstacles, our method outperforms previous state-of-the-art methods in unseen obstacle layouts. In addition, our method demonstrates better generalization w.r.t. different human and obstacle densities.
    \item We successfully transfer the robot policy learned in a low-fidelity simulator to challenging real-world, everyday dynamic environments such as narrow corridors and densely packed public spaces.
\end{enumerate}
 


%% file: Sections/02-LitReview.tex
\section{Related Works}
\label{sec:related}

In this section, we first review the literature of robot crowd navigation (Sec.~\ref{sec:related_model-based} and \ref{sec:related_learning-based}). Then, we narrow down to crowd navigation efforts in constrained spaces (Sec.~\ref{sec:related_constrained}). 
Finally, we review graph attention mechanism with a focus on its usage in multi-agent interaction modeling (Sec.~\ref{sec:related_attn}).

\subsection{Model-based methods}
\label{sec:related_model-based}
Robot navigation in human crowds is particularly challenging and has been studied for decades~\cite{fox1997dynamic,hoy2015algorithms, savkin2014seeking,mavrogiannis2021core}.
Model-based approaches have explored various mathematical models to optimize robot actions~\cite{van2011reciprocal, van2008reciprocal,fox1997dynamic,helbing1995social,huber2022avoiding,mavrogiannis2023winding}. 
As an early example, ROS navigation stack~\cite{ros_navigation_stack} uses a cost map for global planning and dynamic window approach (DWA) for local planning~\cite{fox1997dynamic}. 
DWA searches for the optimal velocity that brings the robot to the goal with the maximum clearance from any obstacle while obeying the dynamics of the robot.
By treating humans as obstacles, DWA exhibits myopic behaviors such as oscillatory paths in dynamic environments~\cite{dobrevski2024dynamic}. As a step forward, optimal reciprocal collision avoidance (ORCA) and social force account for the velocities of agents.
ORCA models other agents as velocity obstacles and assumes that agents avoid each other under the reciprocal rule~\cite{van2011reciprocal, van2008reciprocal}. 
SF models the interactions between the robot and other agents using attractive and repulsive forces~\cite{helbing1995social}.
However, the hyperparameters of the model-based approaches are sensitive to crowd behaviors and thus need to be tuned carefully to ensure good performance~\cite{long2018towards,dobrevski2024dynamic}. In addition, model-based methods are prone to failures, such as the freezing problem, if the assumptions such as the reciprocal rule are broken~\cite{trautman2010unfreezing, liu2024relaxing}.
In contrast, while our method also models these interactions, we learn the hyperparameters of the model from trial and error with minimal assumptions on human behaviors posed by model-based methods. 

\subsection{Learning-based methods}
\label{sec:related_learning-based}
Learning-based approaches have been widely used for navigation in dynamic environments to reduce hyperparameter tuning efforts and the number of assumptions.
One example is supervised learning from expert demonstrations of desired behaviors, where the expert ranges from model-based policies~\cite{long2017deep,tai2018socially,xie2021towards}, human teleoperators in simulators~\cite{pokle2019deep}, to real pedestrians~\cite{karnan2022scand,chandra2024towards}.
Supervised learning does not require explorations of the state and action spaces of the robot, yet the performance of learned policy is limited by the quality of expert demonstrations.

Another line of work takes advantage of crowd simulators and learns policies with RL. Through trial and error, RL has the potential to learn robot behaviors that outperform model-based approaches and expert demonstrations~\cite{xie2023drlvo}. 
For example, Deep V-Learning first uses supervised learning with ORCA as the expert and then uses RL to learn a value function for path planning~\cite{chen2017decentralized,everett2018motion,chen2019crowd,chen2020robot_gaze,yang2023st2}. However, Deep V-Learning assumes that state transitions of all agents are known without uncertainty. In addition, since the networks are pre-trained with supervised
learning, they share the same disadvantages with OCRA, which are hard to correct by RL~\cite{liu2020decentralized}. 
To address these problems, more recent efforts leverage model-free RL without supervised learning or assumptions on state transitions~\cite{liu2020decentralized,mun2023occlusion,liu2024relaxing,yang2023st2}. 
Since the state transition probability of humans is uncertain, directly learning a policy network without explicitly modeling state transitions is more suitable for navigation~\cite{sathyamoorthy2020densecavoid}.
However, the aforementioned RL works have at least one of the following two problems: (1) They focus on navigation in open spaces and isolate agents from static constraints, posing difficulties when deploying robots in the real-world; (2) They ignore all or part of interactions among entities in the scene, which are important for robot navigation in dense crowds and highly constrained environments.
We discuss how prior works address the above two problems in Sec.~\ref{sec:related_constrained} and Sec.~\ref{sec:related_attn} respectively.

\subsection{Scene representation for constrained environments}
\label{sec:related_constrained}
 
To deal with both static constraints and humans, some methods such as DWA~\cite{fox1997dynamic} and DS-RNN~\cite{liu2020decentralized,chen2024decentralized} use groups of small circles to represent the contours of both static obstacles and humans. 
While straightforward, groups of circle representation has the following problems: 1. Humans and obstacles are treated in the same way but have different effects on robot decision making. This design choice leads to suboptimal policy as we will show in Sec.~\ref{sec:sim_exp} and \ref{sec:real_exp}. 2. Since the length of input vectors is proportional to the number of obstacles and humans, the hyperparameters of navigation algorithms easily overfit to a specific environment setting, causing performance degradation when the densities of obstacles and humans change.

Other learning-based approaches use raw sensor images or point clouds to represent the whole environment~\cite{perez2021robot,dugas2022navdreams,pokle2019deep,zheng2022hierarchical,sathyamoorthy2020densecavoid}. 
These end-to-end (e2e) pipelines have made promising progress in simulation.  
However, generalizing these e2e methods to real-world scenarios requires high-fidelity simulators~\cite{tsoi2022sean2,puig2024habitat,Kstner2024Arena3A} and/or datasets~\cite{martin2021jrdb,karnan2022scand,hirose2023sacson}. Since accurate pedestrian simulation is technically difficult and expensive~\cite{Kstner2024Arena3A}, learning a deployable e2e policy in dense and interactive crowds remains an open challenge. 
To circumvent this problem, other works leverage processed state information such as occupancy maps~\cite{xie2023drlvo,yang2023rmrl}.
However, compared to human circles or point clouds, occupancy maps expand the state space, which reduces training efficiency. In addition, the discretization of agent positions limits the ability of occupancy maps to capture slow agent movements.
For this reason, we develop a structured input representation, consisting of human circles and processed obstacle point cloud, that only includes the most essential information for navigation. This input representation leads to strong performance in both simulation and real-world while only requiring a low-fidelity simulator.

\subsection{Graph attention networks for multi-agent interactions}
\label{sec:related_attn}
In recent years, attention mechanisms have demonstrated success in various fields~\cite{vaswani2017attention,velickovic2018graph,vemula2018social}.
Vaswani et al. propose a transformer with self-attention mechanism that achieves state-of-the-art performance in machine translation~\cite{vaswani2017attention}. 
Later, graph attention networks show the effectiveness of attention on learning relationships of data with graphical structures such as citation networks~\cite{velickovic2018graph}.
Inspired by these works, researchers in trajectory prediction and crowd navigation have found that attention networks are also well-suited to capture interactions amongst agents and entities, which contain essential information for multi-agent tasks~\cite{vemula2018social,huang2019stgat, chen2019crowd, liu2020decentralized,hasan2022meta}.
For each agent, these works compute attention scores for all neighboring agents.
Due to the permutation invariance property, attention scores better capture the importance of pair-wise relationships than encoding the states of all agents with a concatenation operation or an recurrent network~\cite{chen2017decentralized,everett2018motion,chen2024decentralized}. 

More specifically, a line of works uses a robot-human attention network to determine the relative importance of each human to the robot~\cite{chen2019crowd, leurent2019social, liu2020decentralized,liu2021learning}.  
However, interactions among humans, which can also influence the robot, are not explicitly modeled.
To this end, another line of works use a homogenous graph network to include both RH and HH interactions~\cite{chen2019relational,yang2023st2, liu2024sample}.
However, since these works feed RH and HH features to a single attention network, the resulting robot policy has difficulty reasoning the specificity of each type of feature, which limits the robot's ability to adapt to different interactions, as demonstrated in \cite{chen2024socially} and our experiments in Sec.~\ref{sec:sim_exp}. 
To deal with this issue, some recent works use multiple attention networks to encode RH and HH features separately, leading to improved navigation performance in dense and interactive crowds~\cite{liu2023intention,wang2023navistar,yang2023st2}.
However, these works only deal with open-world social navigation and thus ignore the interactions between agents and obstacles, which are common in real-world navigation scenarios.

To handle obstacles, Liu et al.~\cite{liu2023graph} and 
Chen et al.~\cite{chen2024socially} treat the humans, the robot, and static objects as nodes in graph neural networks to model the spatio-temporal interactions among all observed entities. 
Different from Chen et al.~\cite{chen2024socially} where a robot needs to reach an object such as a chair, the robot in this work only needs to avoid all objects. Thus, Chen et al.~\cite{chen2024socially} represent static obstacles as a semantic map with a class label for each object, whereas we represent all observed static obstacles as a point cloud. 
In Liu et al.~\cite{liu2023graph}, representing each obstacle as a geometric shape requires further map processing, such as obstacle segmentation. Thus, we treat the point cloud consisting of all obstacles as a single node in our st-graph, which reduces both the processing requirement and the size of the graph.



%% file: Sections/03-Preliminaries.tex
\begin{figure*}[ht]
\centering
\includegraphics[width=\linewidth]{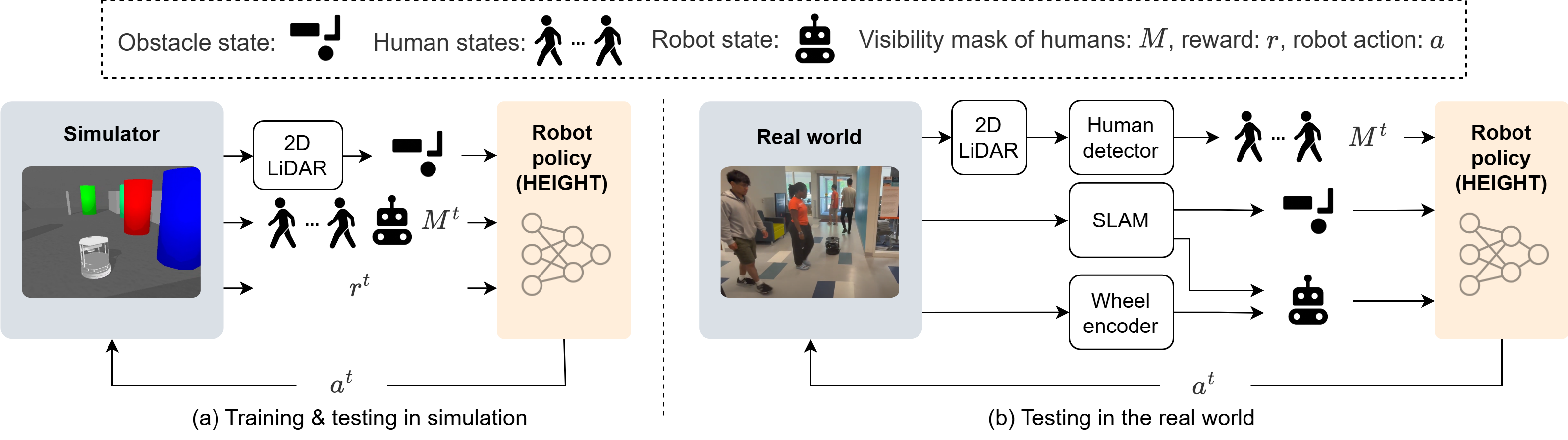}
\caption{\textbf{An overview of our pipeline in simulation and real-world.} (a) At each timestep $t$ in training and testing, the simulator provides a reward $r^t$ and the following observations of the environment: obstacle point cloud $o^t$, the robot state $w^t$, and the human states $h_1^t, ..., h_n^t$, and masks $M^t$ (Sec.~\ref{sec:method_attn}). These observations serve as inputs to \modelName, which outputs a robot action $a^t$ that maximizes the future expected return $R^t$. The simulator executes the actions of all agents and the loop continues. (b) The testing loop in the real-world is similar to the simulator except the perception modules for obtaining the observations are different and the reward is absent. }
\label{fig:pipeline}
\vspace{-15pt}
\end{figure*}

\section{Preliminaries}
\label{sec:prelim}
In this section, we present our problem formulation, scene representation, and reward function.
\subsection{Problem formulation}
\label{sec:prelim-prob}
We model the constrained crowd navigation scenario as a Markov Decision Process (MDP), defined by the tuple $ \langle \mathcal{S}, \mathcal{A}, \mathcal{P}, R, \gamma, \mathcal{S}_0 \rangle$. 
Let $w^t$ be the robot state which consists of the robot's position $(p_x, p_y)$, instantaneous velocity $(u_x, u_y)$, goal position $(g_x, g_y)$, and heading angle $\theta$ in world frame. Let ${h}_i^t$ be the current state of the $i$-th detected human at time $t$, which consists of the human's relative position w.r.t. the robot and its instantaneous velocity $(p_x^i - p_x, p_y^i - p_y, u_x^i, u_y^i)$ in world frame in simulation experiments. In real world experiments, the state of the $i$-th human only includes its relative position $h^t_i=(p_x^i - p_x, p_y^i - p_y)$ since accurate estimation of human velocities are difficult to obtain.
Let ${o}^t$ be the current observation of the static obstacles and walls, which is represented as a 2D point cloud. We define the state $s^t \in \mathcal{S}$ of the MDP as $s^t=[{w}^t, {o}^t, {h}^t_1, ..., {h}^t_n]$ if a total number of $n$ humans are detected at the timestep $t$, where $n$ may change within a range in different timesteps. The IDs of humans are untracked, and their states are ordered by an increasing $L_2$ distance to the robot. 
We add a small Gaussian noise to the ground truth positions and velocities of all agents to compensate for perception errors.
The state space $\mathcal{S}$ is continuous and is expanded in Sec.~\ref{sec:state_rep}. 

In each episode, the robot begins at an initial state $s^0\in \mathcal{S}_0$. As shown in Fig.~\ref{fig:pipeline}(a), according to its policy $\pi(a^t|s^t)$, the robot takes an action $a^t\in\mathcal{A}$ at each timestep $t$. 
The action of the robot consists of the desired translational and rotational accelerations $a^t = [a_{trans}^t, a_{rot}^t]$. The action space $\mathcal{A}$ is discrete: the translational acceleration $a_{trans}\in\{-0.05\,m/s^2, 0\,m/s^2, 0.05\,m/s^2\}$ and the rotational acceleration $a_{rot}\in\{-0.1\,rad/s^2, 0\,rad/s^2, 0.1\,rad/s^2\}$. 
The robot translational velocity is $v_{trans}^t=v_{trans}^{t-1}+a_{trans}^t\times \Delta t \in [-0.5\, m/s, 0.5\, m/s]$ and rotational velocity is $v_{rot}^t=v_{rot}^{t-1}+a_{rot}^t\times \Delta t \in [-1\, rad/s, 1\,rad/s]$.
The robot motion is governed by differential drive dynamics. 
In return, the robot receives a reward $r^t$ (see Sec.~\ref{sec:reward} for details) and transits to the next state $s^{t+1}$ according to an unknown state transition $\mathcal{P}(\cdot|s^t, a^t)$. 
Meanwhile, all other humans also take actions according to their policies and move to the next states with unknown state transition probabilities. 
The process continues until the robot reaches its goal, $t$ exceeds the maximum episode length $T=491$ steps, or the robot collides with any humans or static obstacles. 

The goal of the robot is to maximize the expected return, $R^t=\mathbb{E}[\sum^T_{i=t}\gamma^{i-t}r^{i}]$, where $\gamma$ is a discount factor. The value function $V^\pi (s)$ is defined as the expected return starting from $s$, and successively following policy $\pi$.

\subsection{Scene representation}
\label{sec:state_rep}
\begin{figure}[t]
\centering
\includegraphics[scale=0.4]{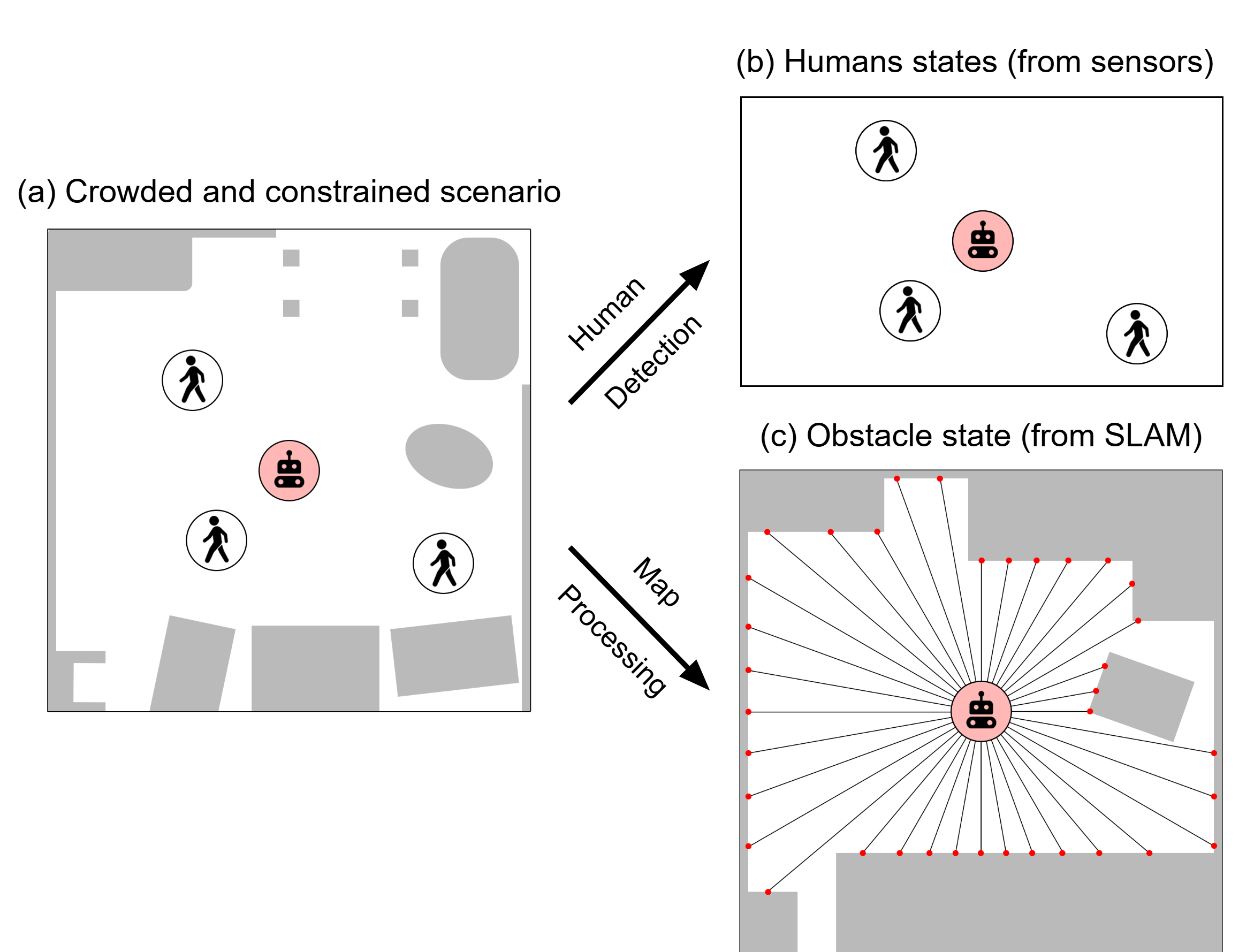}
\caption{\textbf{A split representation of a crowded and constrained navigation scenario.} 
In (b), individual human states are represented by circles. 
To represent static obstacles, we preprocess and smooth the original map (a). 
From this processed map and the robot's localization, we generate an artificial point cloud (c) that simulates obstacle perception. 
}
\label{fig:scene_rep}
\vspace{-15pt}
\end{figure}

To aid policy learning with the complicated state space involving multiple agents and entities, we develop a structured representation that splits human and obstacle states to account for their different effects on robot decision making.
In Fig.~\ref{fig:scene_rep}, at each timestep $t$, we split a scene into a human representation ${h}^t_1, ..., {h}^t_n$, and an obstacle representation ${o}^t$. 
Each human has its own state since it can move independently from one another, whereas all obstacles are mixed together since they are all static.
In human representation (Fig.~\ref{fig:scene_rep}(b)), the position and velocity of each human are detected using off-the-shelf human detectors~\cite{Jia2020DRSPAAM,yolov3,deepsort}. Tracking each human is not required, further simplifying perception requirements. By representing each human as a low-dimentional state vector, we abstract away detailed information such as gaits and appearance, which are difficult to simulate accurately~\cite{tsoi2022sean2,mavrogiannis2023core}. 

We use a 2D point cloud as the obstacle representation, which offers the following two variants. 
The first variant uses a real-time 2D point cloud obtained from onboard sensors. The second variant constructs an ``artificial'' point cloud which is computed from a pre-built map and the robot’s localization in the map. In this case, the point cloud is generated by raycasting from the robot’s estimated pose on the map, representing only static obstacles and unaffected by humans.
The first variant reflects the robot’s instantaneous perception of its surroundings but requires a high-fidelity simulation of LiDAR data. The second variant is more robust to occlusions, imperfect human models, and sensor noise, but assumes the availability of a map and reliable localization.
In this work, since we have access to reliable robot localization and a low-fidelity simulator with inaccurate obstacle shapes and LiDAR point clouds, we opt for the second choice to reduce sim2real gap and improve the poilcy's real-world performance.


\subsection{Reward function}
\label{sec:reward}
Our reward function consists of three parts.
The first and main part of the function awards the robot for reaching its goal and penalizes the robot for colliding with or getting too close to humans or obstacles. In addition, we add a potential-based reward to guide the robot to approach the goal:
\begin{equation}
\label{eq:reward}
\begin{split}
\begin{gathered}
    r_{main}(s^t, a^t)  = 
        \begin{cases}
            20, & \text{if } d_{goal}^t \leq \rho_{robot}\\
            -20, & \text{else if } d_{min}^t \leq 0\\
            d_{min}^t - 0.25, & \text{else if } 0<d_{min}^t<0.25\\
            4(d_{goal}^{t-1}-d_{goal}^t), & \text{otherwise}.
        \end{cases}
\end{gathered}
\end{split}
\end{equation}
where $\rho_{robot}$ is the radius of the robot, $d_{goal}^t$ is the $L2$ distance between the robot and its goal at time $t$, and $d_{min}^t$ is the minimum distance between the robot and any human or obstacle at time $t$. 
The second part is a spinning penalty $r_{spin}$ to penalize high rotational velocity:
\begin{equation}
    r_{spin}(s^t, a^t) = -0.05||\omega^t||_2^2
\end{equation}
where $\omega_t$ is the current rotational velocity of the robot.
Finally, we add another small constant penalty $r_{time}=-0.025$ at each timestep to encourage the robot to finish the episode as soon as possible.

The robot's reward function is the sum of the three parts:
\begin{equation}
    r(s^t, a^t) = r_{main}(s^t, a^t) + r_{spin}(s^t, a^t) +r_{time}
\end{equation}
Intuitively, the robot gets a high reward when it approaches the goal with a high speed and a short and smooth path, while maintaining a safe distance from dynamic and static obstacles. 
The coefficient of each reward term indicates the priority of each subtask (e.x. reaching the goal and avoiding collisions are more important than others) and can be tuned if the desired priority changes.

%% file: Sections/04-Methods.tex
\begin{figure*}[ht]
\centering
\includegraphics[width=\linewidth]{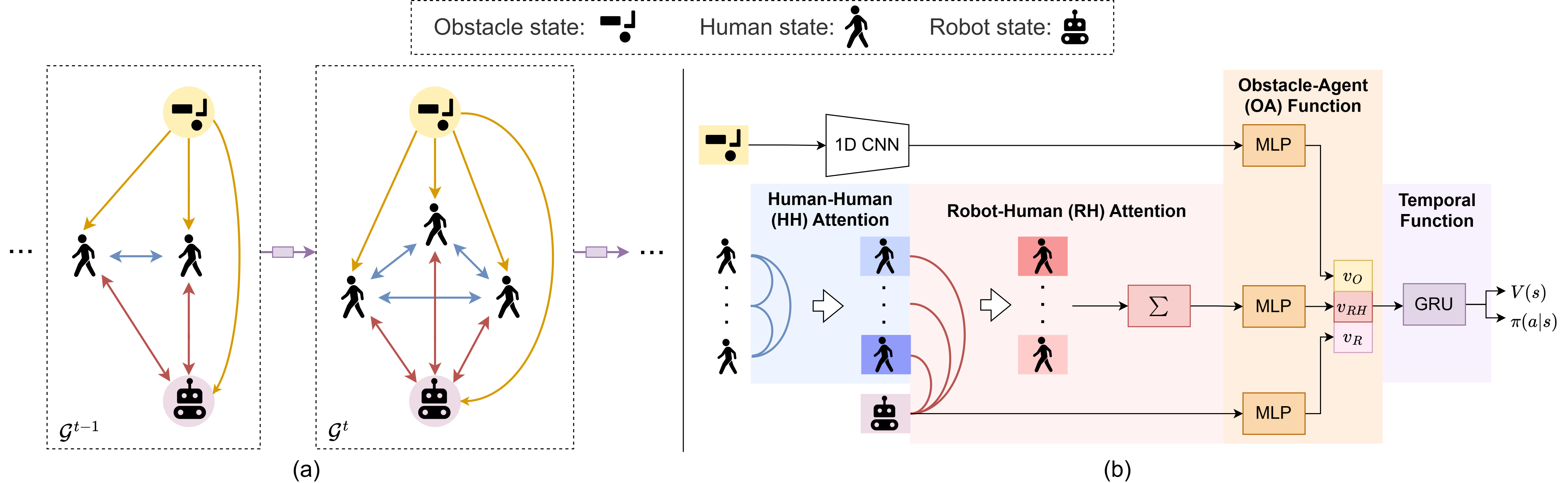}
 \caption{\textbf{The \hstGraph and the \modelName network architecture.} (a) Graph representation of crowd navigation. The robot node is $w$ (pink), the $i$-th human node is $\mathrm{h}_i$ (white), and the obstacle node is $o$ (yellow). HH edges and HH functions are in \textcolor{blue}{blue}, OA edges and OA functions are in \textcolor{YellowOrange}{orange}, and RH edges and RH functions are in \textcolor{red}{red}. The temporal function is in \textcolor{Plum}{purple}. (b) \modelName network. Two attention mechanisms are used to model the HH and RH interactions. We use MLPs and a concatenation for obstacle-agent interactions, and a GRU for the temporal function. The superscript $t$ that indicates the timestep and the human mask $M$ is eliminated for clarity.}
\label{fig:st}
\vspace{-15pt}
\end{figure*}

\section{Methodology}
In this section, we present our approach to decompose the constrained crowd navigation scenario as a heterogeneous st-graph, which leads to the derivation of the \modelName architecture in a structured manner. 

\label{sec:methods}

\subsection{Heterogeneous Spatio-Temporal Graph}
\label{sec:st_graph}
The subtle yet highly dynamic interactions among agents and entities are important factors that makes crowd navigation difficult.
To model these interactions in a structured manner, we formulate the navigation scenario as a heterogeneous st-graph. 
In Fig.~\ref{fig:st}a, at each timestep $t$, our \hstGraph $\mathcal{G}^t = (\mathcal{V}^t, \mathcal{E}^t)$ consists of a set of nodes $\mathcal{V}^t$ and a set of edges $\mathcal{E}^t$. 
The nodes include the detected humans $h_1^t, ..., h_n^t$ and the robot $w^t$. In addition, an obstacle node $o^t$ represents the point cloud of all obstacles as a whole. 
At each timestep $t$, the spatial edges that connect different nodes denote the spatial interactions among nodes. 
Different spatial interactions have different effects on robot decision-making. 
Specifically, since we have control of the robot but not the humans, RH interactions have direct effects while HH interactions have indirect effects on the robot actions. 
As an example of indirect effects, if human A aggressively forces human B to turn toward the robot's front, the robot has to respond as a result of the interaction between A and B.
Additionally, since the agents are dynamic but the obstacles are static, interactions among agents are mutual while the influence of static obstacles on agents is one-way.
Thus, we categorize the spatial edges into three types: HH edges (blue in Fig.~\ref{fig:st}), obstacle-agent (OA) edges (orange), and RH edges (red). 
The three types of edges allow us to factorize the spatial interactions into HH, OA, and RH functions. Each function is parameterized by a neural network that has learnable parameters.
Compared to previous works that ignore some edges~\cite{chen2019crowd,liu2020decentralized,liu2023intention}, our method allows the robot to reason about all observed spatial interactions that exist in constrained and crowded environments.


Since the movements of all agents cause the visibility of each human to change dynamically, the set of nodes $\mathcal{V}^t$ and edges $\mathcal{E}^t$ and the parameters of the interaction functions may change correspondingly.
To this end, we integrate the temporal correlations of the graph $\mathcal{G}^t$ at different timesteps using another function denoted by the purple box in Fig.~\ref{fig:st}(a). The temporal function connects the graphs at adjacent timesteps, which overcomes the short-sightedness of reactive policies and enables long-term decision-making of the robot.

To reduce the number of parameters, the same type of edges in Fig.~\ref{fig:st}(a) share the same function parameters. This parameter sharing is important for the scalability of our graph because the number of parameters is kept constant when the number of human changes~\cite{jain2016structural}.

\subsection{\modelNameUpper Architecture}
In Fig.~\ref{fig:st}b, we derive our network architecture from the \hstGraph. We represent the HH and RH functions as feedforward networks with attention mechanisms, referred to as HH attn and RH attn respectively. We represent the OA function as an MLP with concatenation, and the temporal function as a gated recurrent unit (GRU). We use $W$ and $f$ to denote trainable weights and fully connected layers.

\subsubsection{Attention among agents}
\label{sec:method_attn}
The attention modules assign weights to all edges that connect to a robot or a human node, allowing the node to pay attention to important edges or interactions. The two attention networks are similar to the scaled dot-product attention with a padding mask~\cite{vaswani2017attention}, which computes the attention score using a query $Q$ and a key $K$, and applies the normalized score to a value $V$, which results in a weighted value $v$. 

\begin{equation} \label{eq:single_head_attn}
    v := \textrm{Attn}(Q, K, V, M) = \textrm{softmax}\left(\frac{(QK^\top+M)}{\sqrt{d}}\right)V
\end{equation}
where $d$ is the dimension of the queries and keys and acts as a scaling factor. The mask $M$ is used to handle the changing number of detected humans at each timestep, as we will expand below.

\textbf{Human-human attention:}
To learn the importance of each HH edge to the robot decision at time $t$, we first weigh each observed human w.r.t. other humans using an HH attention network, which is a self-attention among humans. 
In HH attention, the current states of humans $\mathbf{h}^{t}_1, ...,\mathbf{h}^{t}_n$ are concatenated and passed through linear layers with weights $W_{HH}^{Q}$, $W_{HH}^{K}$, and $W_{HH}^{V}$ to obtain $Q_{HH}^t ,K_{HH}^t, V_{HH}^t\in \mathbb{R}^{n \times d_{HH}}$, where $d_{HH}$ is the attention size for the HH attention.
\begin{equation}
    \begin{split}
        Q_{HH}^t = [\mathbf{h}^{t}_1, ...,\mathbf{h}^{t}_n]^\top W_{HH}^{Q} = [q_1, ..., q_n]^\top\\
        K_{HH}^t = [\mathbf{h}^{t}_1, ..., \mathbf{h}^{t}_n]^\top W_{HH}^{K} = [k_1, ..., k_n]^\top\\
        V_{HH}^t = [\mathbf{h}^{t}_1, ..., \mathbf{h}^{t}_n]^\top W_{HH}^{V} = [v_1, ..., v_n]^\top
    \end{split}
\end{equation}
where $q_i\in\mathbb{R}^{1 \times d_{HH}}$, $k_i\in\mathbb{R}^{1 \times d_{HH}}$, and $v_i\in\mathbb{R}^{1 \times d_{HH}}$ are the query embedding, key embedding, and value embedding of the $i$-th human, respectively.

In addition, following Eq.~\ref{eq:single_head_attn}, a mask $M^t\in\mathbb{R}^{n \times n}$ indicates the visibility of each human to the robot at current time $t$ and masks out humans that are undetectable by the robot sensors. 
Assume the $n$-th human is not visible at time $t$. Then $M^t$ is a matrix filled with zeros, except that every entry in $n$-th column is $-\infty$. The numerator of Eq.~\ref{eq:single_head_attn} can be express as
\begin{equation}
\begin{split}
    & \quad\;\; Q_{HH}^t \cdot {\left(K_{HH}^t\right)}^\top  + M^t \\
    &= 
    \begin{bmatrix}
    q_1 k_1 & \cdots & q_1 k_{n-1} & q_1 k_n \\
    \vdots & \ddots & \vdots & \vdots \\
    q_n k_1 & \cdots & q_n k_{n-1} & q_n k_n \\
    \end{bmatrix}
    +
    \begin{bmatrix}
    0 & \cdots & 0 & -\infty \\
    \vdots & \ddots & \vdots & \vdots \\
    0 & \cdots & 0 & -\infty \\
    \end{bmatrix} \\
    &= 
    \begin{bmatrix}
    q_1 k_1 & \cdots & q_1 k_{n-1} & -\infty \\
    \vdots & \ddots & \vdots & \vdots \\
    q_n k_1 & \cdots & q_n k_{n-1} & -\infty \\
    \end{bmatrix}
\end{split}
\end{equation}
Let $V_{HH}^t = [v_1, ..., v_n]^\top$, where $v_i\in\mathbb{R}^{1 \times d_{HH}}$ is the value embedding of the $i$-th human. Then, based on Eq.~\ref{eq:single_head_attn}, the weighted human embeddings $v_{HH}^t\in \mathbb{R}^{n \times d_{HH}}$ is 
\begin{equation}\label{eq:softmax_term}
\begin{split}
    v_{HH}^t
    &= \textrm{softmax}\left(\frac{Q_{HH}^t {\left(K_{HH}^t\right)}^\top+M^t}{\sqrt{d}}\right) \cdot V_{HH}^t \\
    &= 
    \begin{bmatrix}
    c_1 q_1 k_1 & \cdots & c_1 q_1 k_{n-1} & 0 \\
    \vdots & \ddots & \vdots & \vdots \\
    c_n q_n k_1 & \cdots & c_n q_n k_{n-1} & 0 \\
    \end{bmatrix}
    \cdot
    \begin{bmatrix}
    v_1 \\
    \vdots \\
    v_{n-1} \\
    v_n \\
    \end{bmatrix} \\
    &= 
    \begin{bmatrix}
     c_1 q_1 k_1 v_1 + \cdots +  c_1 q_1 k_{n-1} v_{n-1} \\
    \vdots \\
    c_n q_n k_1 v_1 + \cdots + c_n q_n k_{n-1} v_{n-1} \\
    \end{bmatrix}
\end{split}
\end{equation}
where $c_1, ..., c_n$ are constants that reflect the effect of the scaling factor $d$ and the softmax function.
Thus, the value of the $n$-th missing human $v_n$ is eliminated by the mask $M^t$ and thus does not affect the resulting weighted human embedding $v_{HH}^t$. 
The mask that indicates the visibility of each human prevents attention to undetected humans, which is common in crowd navigation due to the partial observability caused by the limited robot sensor range, occlusions, imperfect human detectors, etc~\cite{mun2023occlusion}. Additionally, the mask provides unbiased gradients to the networks, which stabilizes and accelerates the training~\cite{ma2020reinforcement,liu2023intention}.


\textbf{Robot-human attention:}
After the humans are weighted by HH attention, we weigh their embeddings again w.r.t. the robot with another RH attention network to learn the importance of each RH edge. 
In RH attention, we first embed the robot state ${w}^t$ with a fully connected layer, which results in the key for RH attention $K_{RH}^t\in \mathbb{R}^{1 \times d_{RH}}$. 
The query and the value, $Q_{RH}^t, V_{RH}^t\in \mathbb{R}^{n \times d_{RH}}$, are the other two linear embeddings of the weighted human features from HH attention $v_{HH}^t$. 
\begin{equation}
    \begin{split}
Q_{RH}^t = v_{HH}^t W_{RH}^{Q},\:
K_{RH}^t = {w}^t W_{RH}^{K},\:
V_{RH}^t = v_{HH}^t W_{RH}^{V}
 \end{split}
\end{equation}
We compute the attention score from $Q_{RH}^t$, $K_{RH}^t$, $V_{RH}^t$, and the mask $M^t$ to obtain the weighted human features for the second time $v_{RH}^t\in \mathbb{R}^{1 \times d_{RH}}$ as in Eq.~\ref{eq:single_head_attn}. 


\subsubsection{Incoporating obstacle and temporal information}
We first feed the point cloud that represents obstacles, $o^t$, into a 1D-CNN followed by a fully connected layer to get an obstacle embedding $v^t_O$. Then, we embed the robot states ${w}^t$ with linear layers $f_{R}$ to obtain a robot embedding $v_R^t$.
\begin{equation}
   v^t_O = f_{CNN}(o^t),\quad v_R^t=f_{R}({w}^t) 
\end{equation}
Finally, the robot and obstacle embeddings are concatenated with the twice weighted human features $v_{RH}^t$ and fed into the GRU\footnote{From experiments, we find that adding a third obstacle-agent attention network, where the obstacle embedding is the key and the robot and human embeddings are the query and value, does not improve the performance. The reason is that the agents only interact with a small subset of obstacles instead of the whole point cloud at each timestep, eliminating the need for modeling these interactions.} Thus, we keep the simple concatenation design to incorporate obstacle information.:
\begin{equation}
	 h^t=\mathrm{GRU}\left(h^{t-1}, ([v_{RH}^t, v_R^t, v^t_O])\right)
\end{equation}
where $h^t$ is the hidden state of GRU at time $t$.
Finally, the $h^t$ is input to a fully connected layer to obtain the value $V(s^t)$ and the policy $\pi(a^t|s^t)$.

\subsubsection{Training}
We train the entire network with Proximal Policy Optimization (PPO)~\cite{schulman2017proximal} in a simulator as shown in Fig.~\ref{fig:pipeline}(a) (see Sec.~\ref{sec:sim_env} for details of the simulator). At each timestep $t$, the simulator provides all state information that constitutes $s^t$, which is fed to the \modelName network. The network outputs the estimated value of the state $V(s^t)$ and the logarithmic probabilities of the robot's action $\pi(a^t|s^t)$, both of which are used to compute the PPO loss and then update the parameters in the network. 
In training, the robot action is sampled from the action distribution $\pi(a^t|s^t)$. In testing, the robot takes the action with the highest probability $a^t$. The robot action $a^t$ is fed into the simulator to compute the next state $s^{t+1}$, and then the loop continues. 

Without any supervised learning, our method is not limited by the performance of expert demonstrations~\cite{chen2017decentralized,chen2019crowd,liu2024sample}. However, to improve the low training data efficiency, an inherent problem of RL, \modelName can also be trained with a combination of imitation learning and RL. 


\subsubsection{Summary}
We present a structured and principled approach to design the robot policy network for crowd navigation in constrained environments. 
By decomposing the complex scenario into independent components, we split the complex problem into smaller functions, which are used to learn the parameters
of the corresponding functions. By combining all components
above, the end-to-end trainable \modelName allows the robot to perform
spatial and temporal reasoning on all pairwise interactions, leading to better navigation performance.


%% file: Sections/05-SimExp.tex
\section{Simulation Experiments}

\label{sec:sim_exp}
In this section, we present our environment, experiment setup, and results in simulation. 
Our experiments are guided by the following questions: 
(1) What is the advantage of our split scene representation compared with alternative representations?
(2) What is the importance of the graph formulation and why do we differentiate types of edges with a heterogenous st-graph?
(3) What is the importance of HH and RH attention in \modelName?
(4) What are the failure cases of our method?


\subsection{Simulation benchmark}
\label{sec:sim_env} 
We conduct simulation experiments in a PyBullet simulator~\cite{coumans2019}. The robot, humans, and static obstacles are in a 12\,m$\times$12\,m arena. In each episode, rectangular obstacles are initialized with random shapes and random poses. The width and the length of each static obstacle are sampled from $\mathcal{N}(1, 0.6^2)$ and are limited in $[0.1, 5]$ meters. The initial positions of the humans and the robot are also randomized. The starting and goal positions of the robot are randomly sampled inside the arena. The distance between the robot's starting and the goal positions is between 5\,m and 6\,m. 
Some humans are dynamic and some are static.
The goals of dynamic humans are set on the opposite side of their initial positions to create circle-crossing scenarios. In training, the number of humans varies from 5 to 9 and the number of static rectangular obstacles varies from 8 to 12. Among all humans, 0-2 of them are static and the rest are dynamic. In testing, the number of humans and obstacles are shown in the first column of Table~\ref{tab:baseline_results}. The heights of humans and obstacles are all above the height of the robot LiDAR to ensure detectability. 

To simulate a continuous human flow similar to \cite{liu2020decentralized,liu2023intention,xie2023drlvo}, dynamic humans will move to new random goals immediately after they arrive at their goal positions or they get stuck in front of narrow passageways for more than 10 timesteps.
All dynamic humans are controlled by ORCA~\cite{van2011reciprocal}.  80\% of dynamic humans do not react to the robot and 20\% of humans react to the robot. This mixed setting prevents our network from
learning an extremely aggressive policy in which the robot
forces all humans to yield while achieving a high reward. Meanwhile, the simulation maintains enough reactive humans to resemble the real crowd behaviors. 
We use holonomic kinematics for humans. 
The preferred speed of humans ranges from 0.4\,m/s to  0.6\,m/s to accommodate the speed of the robot. 
We assume that humans can achieve the desired velocities immediately, and they will keep moving with these velocities for the next $\Delta t$ seconds.
The simulation and control frequency $\Delta t$ is 0.1\,s.
The maximum length of an episode is 491 timesteps or 49.1\,s. 

\subsection{Setup}


\subsubsection{Baselines}

We compare the performance of our method with classical and state-of-the-art baselines listed in the top 5 row of Table~\ref{tab:baseline_intro}. The baselines vary in terms of both scene representation and navigation algorithm. A detailed description of all baselines can be found in Appendix\ref{sec:append_baseline}.

\begin{table*}[t]
\centering
\caption{Comparison of scene representation and interaction modeling of all methods}
\label{tab:baseline_intro}
\begin{tabular}{p{1.7cm} p{6.5cm} p{8.5cm}}
\toprule
\textbf{Method} & \textbf{Scene Representation (humans and static obstacles)} & \textbf{Navigation algorithm} \\
\midrule
\textbf{DWA}~\cite{fox1997dynamic} & Circles for both & Treats humans as obstacles, no interaction modeling \\
\textbf{ORCA}~\cite{van2011reciprocal} & Circles for humans + Line segments for obstacles & Optimization program with velocity obstacles to model all interactions \\
\textbf{\astarCNN}~\cite{perez2021robot} & LiDAR point cloud for both & A* + RL policy with no interaction modeling  \\
\textbf{DS-RNN}~\cite{liu2020decentralized} & Circles for both & RL policy with only RH attention and GRU \\
\textbf{DRL-VO}~\cite{xie2023drlvo} & Occupancy map for humans + Point cloud for both & Pure pursuit algorithm + RL policy with no interaction modeling \\
\textbf{\homoBaseline}~\cite{velickovic2018graph} & Circles for humans + Point cloud for obstacles & RL policy from st-graph with shared attention weights for all interactions \\
\textbf{Ours} & Circles for humans + Point cloud for obstacles & RL policy from st-graph with separate functions for all interactions \\
\bottomrule
\end{tabular}
\vspace{-10pt}
\end{table*}
 
\subsubsection{Ablations}
To show the benefits of each attention network, we perform an ablation study on the two attention models. The ablated models are defined as follows:
\begin{itemize}
    \item \textbf{No attn}: Both RH and HH attention networks are removed. No attn model only has the embedding layers for the inputs and the GRU.
    \item \textbf{RH}: The HH attention network is removed and the humans are weighted only once w.r.t. the robot. Everything else is the same as ours.
    \item \textbf{HH}: The RH attention network is removed and the humans are weighted only once w.r.t. other humans. Everything else is the same as ours.
    \item \textbf{RH+HH (ours)}: The full network as shown in Fig.~\ref{fig:st}(b).
\end{itemize}

\subsubsection{Training} For fair comparision, we train all RL methods, including all baseines and ablations except DWA, with the same number of timesteps and the same PPO hyperparameters. Please refer to Appendix\ref{sec:append_resource}-1) for training details. 

\subsubsection{Evaluation}
We test all methods with the same $500$ random unseen test episodes. 
For RL-based methods, we test the last 5 checkpoints (equivalent to the last 6000 training steps) and report the result of the checkpoint with the highest success rate.
We conduct the testing in 5 different human and obstacle densities, as shown in the first columns of Table~\ref{tab:baseline_results} and Table~\ref{tab:ablation_results}. The first set of densities is the same as the training environment and is thus in \textit{training distribution}. To test the generalization of methods, we change the range of human or obstacle numbers in the remaining four environments to values that do not overlap with those used in training. Thus, these 4 environments are out-of-distribution (OOD). 

All testing scenarios, including those in the \textit{training distribution}, are unseen during training. This is because in each testing episode, a new seed that is not used in training determines the number of humans and obstacles, the starting and goal positions of all agents, and the size and pose of obstacles. Only the 4 arena walls do not change across different episodes.

The testing metrics measure the quality of the navigation and include the success rate, collision rate with humans and obstacles, timeout rate, the average navigation time of successful episodes in seconds, and path length of the successful episodes in meters.

\begin{table*}[t]
  \begin{center}
    \caption{Baseline comparison results with different human and obstacle densities in unseen environments}
    \label{tab:baseline_results}
    \begin{tabular}{c l c c c c c c c }
    \toprule
     \multirow{2}{*}{\textbf{Environment}} & \multirow{2}{*}{\textbf{Method}} & \multirow{2}{*}{\textbf{Success}$\uparrow$}  &  \multicolumn{3}{c}{\textbf{Collision}$\downarrow$} & \multirow{2}{*}{\textbf{Timeout}$\downarrow$} & \multirow{2}{*}{\textbf{Nav Time}$\downarrow$} & \multirow{2}{*}{\textbf{Path Len}$\downarrow$} \\
     \cmidrule{4-6}
     &&& \textbf{Overall}&\textbf{w/ Humans}&\textbf{w/ Obstacles}& &   \\
     
     \midrule
     \multirow{7}{*}{\shortstack{\textbf{Training distribution} \\ 5-9 humans \\ 8-12 obstacles }} 
    & DWA~\cite{fox1997dynamic} &0.16 &  0.68&0.63 & 0.05 &0.15 &28.45 & 11.52 \\
    & ORCA~\cite{van2011reciprocal} &0.34 &  0.62&0.31 & 0.32 &0.04 &21.51 & 9.46 \\
    & \astarCNN~\cite{perez2021robot} & 0.64 &  0.29&0.28&\textbf{0.01}&0.07&25.72&12.30\\
    & DS-RNN~\cite{liu2020decentralized} &0.80 & 0.20&0.11&0.09& \textbf{0.00}&19.90&10.78\\
    & DRL-VO~\cite{xie2023drlvo}  &0.59 &  0.41& 0.34 & 0.07 & \textbf{0.00} & 21.45 &10.36 \\
    & \homoBaseline~\cite{velickovic2018graph} &0.86 &  0.14&0.13 &\textbf{0.01} & \textbf{0.00} & 18.66 &10.36\\
    & \modelName (ours)  &\textbf{0.88} &  \textbf{0.12}&\textbf{0.09} & 0.03 & \textbf{0.00} & \textbf{18.31} &\textbf{10.34} \\
     
      \midrule
      \multirow{7}{*}{\shortstack{\textbf{Less crowded}  \\0-4 humans \\ 8-12 obstacles}}  & DWA~\cite{fox1997dynamic} & 0.29 &  0.37&0.28&0.09&0.34&25.74 & 14.42 \\
     & ORCA~\cite{van2011reciprocal} & 0.42 &  0.54&0.11&0.42&0.04&19.77 & 9.69 \\
     & \astarCNN~\cite{perez2021robot}&0.84& 0.12&0.11&\textbf{0.01}&0.04&22.64&11.73 \\
     & DS-RNN~\cite{liu2020decentralized}& 0.72 &  0.28&0.07 & 0.21 & \textbf{0.00} & 17.21 & \textbf{9.56} \\
     & DRL-VO~\cite{xie2023drlvo}  &0.82 &  0.17& 0.12 & 0.05 & 0.01 & 19.14 &10.53 \\
     & \homoBaseline~\cite{velickovic2018graph}&0.95 & 0.05&0.03&0.02&\textbf{0.00}&16.64&10.17\\
     & \modelName (ours)&\textbf{0.97} &  \textbf{0.03}&\textbf{0.02} & \textbf{0.01}  & \textbf{0.00} & \textbf{16.32} & 10.04 \\
      \midrule
      \multirow{7}{*}{\shortstack{\textbf{More crowded}  \\10-14 humans \\ 8-12 obstacles }}  & DWA~\cite{fox1997dynamic} &0.11 &  0.78&0.74 & 0.04 & 0.11 & 29.60 & \textbf{10.37} \\
     & ORCA~\cite{van2011reciprocal} &0.26 &  0.71&0.41 & 0.30 & 0.03&23.33 & 9.21 \\
     & \astarCNN~\cite{perez2021robot}&0.47& 0.42&0.39&\textbf{0.03}&0.11&27.33&12.47 \\
     & DS-RNN~\cite{liu2020decentralized}&0.76 &  \textbf{0.22}&\textbf{0.16} & 0.06 & 0.01 & 23.08 & 11.67 \\
     & DRL-VO~\cite{xie2023drlvo}  &0.50 &  0.49& 0.41 & 0.09 & 0.01 &22.72&10.26 \\
     & \homoBaseline~\cite{velickovic2018graph}&0.72& 0.28&0.23&0.05&\textbf{0.00}&20.46&10.57\\
     & \modelName (ours)& \textbf{0.78} &  \textbf{0.22}&0.19 & \textbf{0.03} & \textbf{0.00} & \textbf{19.69} & 10.39 \\
      
      \midrule
      \multirow{7}{*}{\shortstack{\textbf{Less constrained} \\ 5-9 humans \\ 3-7 obstacles }}  & DWA~\cite{fox1997dynamic} &0.17 &  0.57&0.55 & 0.02  & 0.26&28.50 & 14.21 \\
     & ORCA~\cite{van2011reciprocal} &0.42 &  0.54&0.30 & 0.24 & 0.03&21.89 & 10.59 \\
     & \astarCNN~\cite{perez2021robot}&0.66& 0.29&0.28&\textbf{0.01}&0.05&23.63&11.98 \\
     & DS-RNN~\cite{liu2020decentralized}&0.66 &  0.34&0.20 & 0.14 & \textbf{0.00} & 18.01& \textbf{9.76} \\
     & DRL-VO~\cite{xie2023drlvo}  &0.70 &  0.30& 0.26 & 0.04 & 0.00 &21.21&11.15 \\
     & \homoBaseline~\cite{velickovic2018graph}&0.88& 0.12&0.10&0.02&\textbf{0.00}&17.66&10.37 \\
     & \modelName (ours)& \textbf{0.90} &  \textbf{0.10}&\textbf{0.09} & \textbf{0.01} & \textbf{0.00} & \textbf{17.20} & 10.23 \\
      
\midrule
      \multirow{6}{*}{\shortstack{\textbf{More constrained}\\ 5-9 humans \\ 13-17 obstacles}}  & DWA~\cite{fox1997dynamic} &0.14 &  0.66&0.49 & 0.17& 0.20 & 30.47 & 11.35 \\
     & ORCA~\cite{van2011reciprocal} &0.30 &  0.58&0.24 & 0.34 & 0.11&22.40 & 9.04 \\
     & \astarCNN~\cite{perez2021robot}&0.48& 0.29&0.23&\textbf{0.06}&0.23&27.28&13.11\\
     & DS-RNN~\cite{liu2020decentralized}&0.71&  0.23&0.09& 0.14 & 0.06 &23.45 & 11.58 \\
     & DRL-VO~\cite{xie2023drlvo}  &0.55 &  0.40& 0.23 & 0.07 & 0.05 &21.87&10.22 \\
     & \homoBaseline~\cite{velickovic2018graph}&0.80& 0.20&0.11&0.09&\textbf{0.00}&19.20&\textbf{10.51}\\
     & \modelName (ours)& \textbf{0.84} &  \textbf{0.15}&\textbf{0.07 }& 0.08 & 0.01 & \textbf{18.79} & 10.65 \\
      \bottomrule
    \end{tabular}
  \end{center}
  \vspace{-12pt}
\end{table*}

\subsection{Results}
\subsubsection{Effectiveness of scene representation}
\label{sec:result_rep}
To analyze the effects of input scene representations on crowd navigation algorithms, we compare ours and \homoBaseline, the two methods that distinguish human and obstacle inputs, with baselines that mix humans and obstacles in input representation: the model-based DWA, the RL-based DS-RNN, and the hybrid planner \astarCNN. The results are shown in Table~\ref{tab:baseline_results}. 

For DWA, treating humans as obstacles leads to the highest average human collision rates (0.54) and a freezing problem, indicated by the highest average timeout rates (0.21), in all environments. 
For example, the robot in Fig.~\ref{fig:sim_more_constrained} stays close to everything and thus fails to avoid the magenta human in time.


Similarly, by representing obstacles as groups of circles, DS-RNN performs better in the \textit{More crowded} environment (Fig.~\ref{fig:sim_more_crowded}(d)) than in the \textit{More constrained} environment (Fig.~\ref{fig:sim_more_constrained}(d)). In addition, Table~\ref{tab:baseline_results} shows that DS-RNN achieves the highest average collision rate with obstacles (0.13) in all environments. Furthermore, the obstacle collision rate of DS-RNN increases in all 4 OOD environments, indicating an overfitting problem. Among the OOD environment, the percentage of obstacle collision increase is higher in environments with higher obstacle-to-human ratios. For example, in \textit{Less crowded} with the highest percentage of obstacles, the obstacle collision rate (0.21) increases by 133\% w.r.t. the obstacle collision rate in \textit{training distribution} (0.09). On the contrary, in  \textit{More crowded} with the lowest percentage of obstacles, the obstacle collision rate (0.06) drops by 33\%. 
The reason is that DS-RNN has trouble inferring the geometric shapes of obstacles from a large group of circles, and thus fails to avoid collision with them. 
Thus, treating both humans and obstacles as circles is not an ideal input representation for robot navigation algorithms. 

For \astarCNN, the $\textrm{A}^{*}$ global planner does not account for humans and the CNN local planner represents all observations as a 2D point cloud. 
As a result, from Table~\ref{tab:baseline_results}, \astarCNN has the highest average timeout rate (0.10) and the longest average time (18.99) among RL-based methods in all environments, especially the two most challenging ones, for the following 2 reasons. 
(1) The waypoints can fail to guide the robot to reach the goal because the waypoints lose optimality as agents move. 
Consequently, in OOD scenarios such as Fig.~\ref{fig:sim_more_constrained}(b), the CNN policy is especially bad at long-horizon planning because it is overfitted with good waypoints.
(2) By mixing humans and obstacles in its input, the robot policy sometimes has a hard time distinguishing dynamic and static obstacles. For example, in Fig.~\ref{fig:sim_more_constrained}(b), the robot keeps a large distance from everything and thus is less agile and efficient compared with the robots in Fig.~\ref{fig:sim_more_constrained}(e) and (f). 
Thus, for RL-based approaches, treating humans as obstacles leads to overfitting to specific types of scenarios, preventing the policies from generalizing to OOD scenarios which are common in the real-world. 

Compared with \astarCNN, the success rate of DRL-VO is not always better but degrades less in \textit{More crowded, Less constrained,} and \textit{More constrained} scenarios. This difference can be attributed to the input representation of its RL policy, which includes low-level human states encoded as occupancy maps in addition to point cloud data. Since the size of occupancy maps remains fixed regardless of the number of pedestrians, the policy generalizes well across varying human densities. However, occupancy maps have a much higher dimensionality compared to point clouds or detected human circles, which can lead to underfitting under the same training budget as \homoBaseline and our method (Fig.~\ref{fig:sim_more_constrained}(f)).


By splitting human and obstacle input representations, \homoBaseline and \modelName achieve the top 2 performance across most metrics and environments. Especially, \modelName learns to keep a larger distance from human paths yet a shorter distance from obstacles to balance safety and efficiency (Fig.~\ref{fig:sim_more_constrained}(g)), indicated by the lowest overall collision rates (0.17) and the shortest average navigation time (18.10) in all environments.
In addition, in the last 3 frames of Fig.~\ref{fig:sim_more_crowded}(e), the robot detours around the thin and long wall because the wall will not move out of its way. 
However, when the robot encounters the leftmost green human, it waits for the human to pass and resume afterwards because the human only blocks its way temporarily.
Therefore, we conclude that our split representation consisting of detected human states and obstacle point clouds is most suitable to learn robot navigation in crowded and constrained environments with RL, among all popular choices in Table~\ref{tab:baseline_results}.

\subsubsection{Effectiveness of the \hstGraph}
To justify our \hstGraph, we compare our method with ORCA,  \astarCNN, and DRL-VO with no graph concept and \homoBaseline with a homogenous graph in Table~\ref{tab:baseline_results}.

\begin{figure*}

    \centering
    \includegraphics[width=\linewidth]{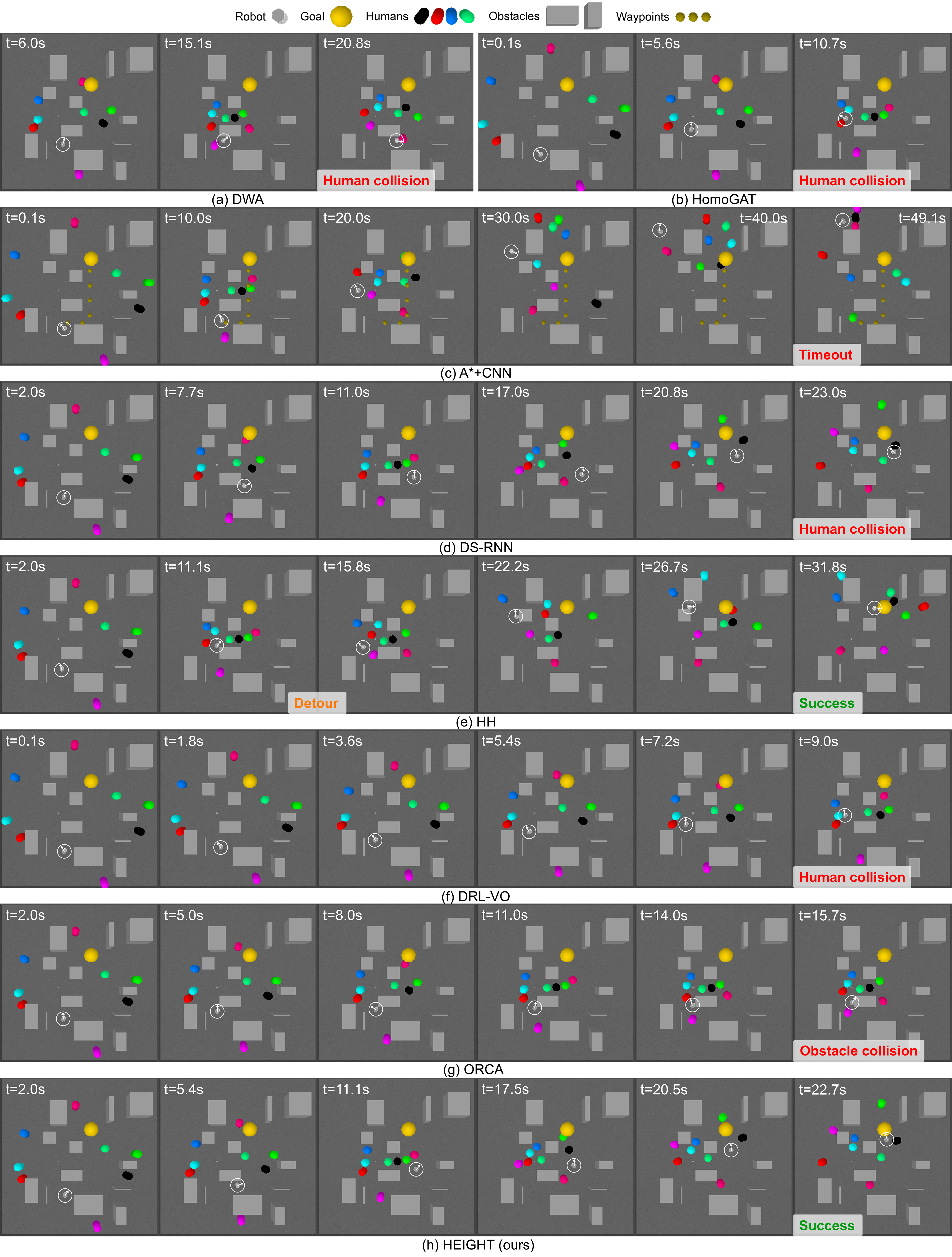}
    
    
    \captionof{figure}{\textbf{Comparison of different methods in the same testing episode in \textit{More Constrained} environment.} 
    The robot is centered in white circles and its orientation is denoted by white arrows. 
    More qualitative results can be found in the video attachment.
    }
    \vspace{-5pt}
    \label{fig:sim_more_constrained}
\end{figure*}

ORCA models the RH and OA interactions at the current timestep as velocity obstacles. As a result, ORCA is the second worst among all methods in terms of success rate and performs especially poorly in terms of obstacle collision rates, shown in Fig.~\ref{fig:sim_more_constrained}(g). Since all methods that outperform ORCA contain either a search-based planner or recurrent networks, this result indicates the importance of HH interaction and long-term temporal reasoning in our problem. 
However, ORCA has the shortest average path length in most environments due to its greedy decisions to move toward the goal.

\astarCNN lacks structure in input representation, and subsequently, lacks structure in network architecture. For this reason, as we discussed in Sec.~\ref{sec:result_rep}, \astarCNN shows much worse average success rate (0.62 v.s. 0.84), navigation time (25.32 v.s. 18.52), and path length (12.32 v.s. 10.40) than \homoBaseline which has a simple st-graph in all environments. \astarCNN also exhibits larger variances in terms of success rate (0.019 v.s. 0.0060), navigation time (3.62 v.s. 1.70), and path length (0.22 v.s. 0.019)  across four OOD environments. 
Especially in the challenging \textit{More crowded} and \textit{More constrained} environments, \astarCNN exhibits a larger drop in average success rate compared with \homoBaseline (0.165 v.s. 0.10).
A similar phenomenon can also be observed by comparing DRL-VO, which only lacks structure in network architecture, and \homoBaseline.
This finding shows that even a simple st-graph structure can lead to a performance gain, indicating the importance of spatial and temporal reasoning for constrained robot crowd navigation.

\begin{figure*}[ht]

    \centering
    \includegraphics[width=\linewidth]{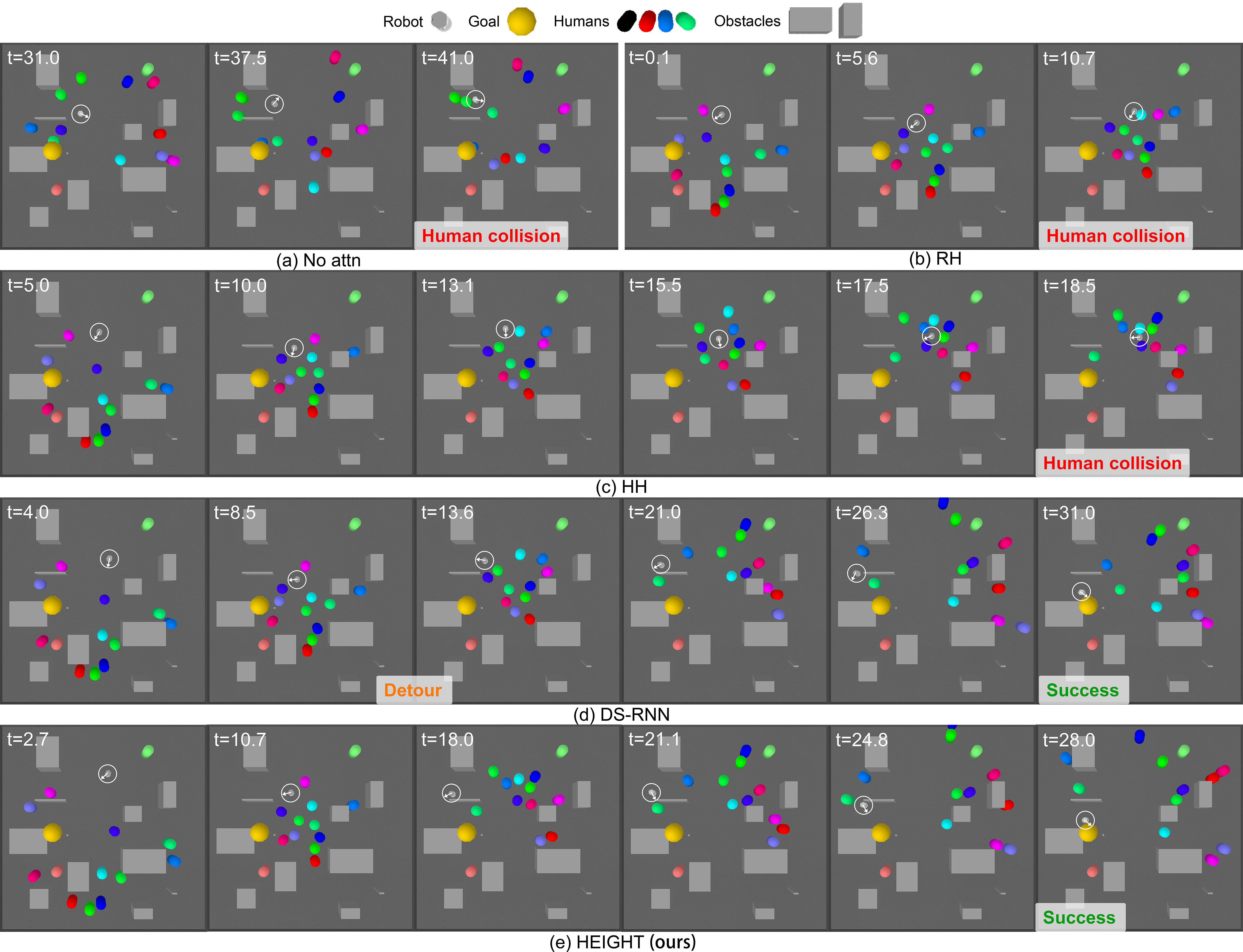}
    
    
    \caption{Comparison of different methods in the same testing episode in \textit{More Crowded} environment.}
    \vspace{-10pt}
    \label{fig:sim_more_crowded}
\end{figure*}

However, occasional failures of \homoBaseline still exist such as Fig.~\ref{fig:sim_more_constrained}(b). Without differentiating RH and OA interactions, the robot maintains similar distances from humans and obstacles, which is prone to collisions because it needs to keep larger distances from humans.
As a step further,  with a \hstGraph, \modelName treats the 3 types of edges with different network components. As a result, \modelName outperforms \homoBaseline and achieves the best average success rate (0.87), navigation time (18.06), and path length (10.33) in all environments. 
\modelName's variance of most metrics across all testing environments are the smallest, indicating its robustness to distribution shift.
The reason is that the heterogeneous components allow the robot to reason about the different effects of HH, RH, and OA spatial interactions. For example, in Fig.~\ref{fig:sim_more_constrained}(f) and Fig.~\ref{fig:sim_more_crowded}(e), \modelName chooses a path that avoids the most crowded region, yields to humans when the robot must encounter them, and stays closer to walls for efficiency. 
Therefore, we conclude that the spatial and temporal reasoning on different types of interactions is the key to ensuring good in-distribution performance and OOD generalization in crowded and interactive environments.

\begin{table*}[t]
  \begin{center}
    \caption{Ablation study results with different human and obstacle densities in unseen environments}
    \label{tab:ablation_results}
    \begin{tabular}{c l c c c c c c c}
    \toprule
     \multirow{2}{*}{\textbf{Environment}} & \multirow{2}{*}{\textbf{Method}} & \multirow{2}{*}{\textbf{Success}$\uparrow$}  & \multicolumn{3}{c}{\textbf{Collision}$\downarrow$} & \multirow{2}{*}{\textbf{Timeout}$\downarrow$} & \multirow{2}{*}{\textbf{Nav Time}$\downarrow$} & \multirow{2}{*}{\textbf{Path Len}$\downarrow$}\\
     \cmidrule{4-6}
     &&&\textbf{Overall}&\textbf{w/ Humans}&\textbf{w/ Obstacles}& & & \\
     
     \midrule
     \multirow{4}{*}{\shortstack{\textbf{Training distribution} \\ 5-9 humans \\ 8-12 obstacles }} 
    & No attn &0.52 &0.46&0.41&0.05&0.02&26.48&11.81 \\
    & RH & 0.85 & 0.15& 0.13 & \textbf{0.02} & \textbf{0.00} & 18.86 & 10.41 \\
    & HH & 0.87  & \textbf{0.12}& 0.10 & \textbf{0.02}& 0.01& \textbf{18.31} & 10.42  \\
    & RH+HH (ours) &\textbf{0.88} & \textbf{0.12}& \textbf{0.08} & 0.04 & \textbf{0.00} & 18.49 &\textbf{10.28} \\

      \midrule
      \multirow{4}{*}{\shortstack{\textbf{Less crowded}  \\0-4 humans \\ 8-12 obstacles}} 
       & No attn &0.72 & 0.27&0.24&0.03&0.01&21.78&10.94 \\
    & RH &0.89 & 0.07& 0.04 & 0.03 & 0.04 & 17.34 & 10.66   \\
    & HH & 0.94 & 0.06& 0.05&\textbf{0.01}&\textbf{0.00}&\textbf{16.10}&\textbf{9.90}  \\
    & RH+HH (ours) &\textbf{0.97} & \textbf{0.03}& \textbf{0.02} & \textbf{0.01}  & \textbf{0.00} & 16.32 & 10.04 \\
    
      \midrule
      \multirow{4}{*}{\shortstack{\textbf{More crowded}  \\10-14 humans \\ 8-12 obstacles }} 
       & No attn &0.29 & 0.61&0.53&0.08&0.10&28.35&12.57  \\
    & RH & 0.70 & 0.29& 0.25 & 0.04 & 0.01 & 20.39 & 10.49  \\
    & HH &0.74 & 0.25& \textbf{0.17} & 0.08 & 0.01& 20.34 & 10.53 \\
    & RH+HH (ours) & \textbf{0.78} & \textbf{0.22}& 0.19 & \textbf{0.03} & \textbf{0.00} & \textbf{19.69} & \textbf{10.39}  \\
      
      \midrule
      \multirow{4}{*}{\shortstack{\textbf{Less constrained} \\ 5-9 humans \\ 3-7 obstacles }} 
       & No attn &0.55 & 0.43&0.42&\textbf{0.01}&0.01&24.64&11.75 \\
    & RH & 0.86  & 0.14& 0.12 & 0.02 & \textbf{0.00} & 18.17& 10.33  \\
    & HH & 0.88 & 0.12& 0.10& 0.02& \textbf{0.00} &17.24&10.25  \\
    & RH+HH (ours)& \textbf{0.90} & \textbf{0.10}& \textbf{0.09} & \textbf{0.01} & \textbf{0.00} & \textbf{17.20} & \textbf{10.23}  \\
      
\midrule
      \multirow{4}{*}{\shortstack{\textbf{More constrained}\\ 5-9 humans \\ 13-17 obstacles}} 
       & No attn &0.43 & 0.49&0.38&0.11&0.08&26.25&11.82 \\
    & RH &0.77  & 0.22& 0.11 & 0.11 & 0.01 & 20.21 & 10.73 \\
    & HH &\textbf{0.84}  & 0.16& \textbf{0.06} & 0.10 & \textbf{0.00} & 18.94 & 10.66  \\

    & RH+HH (ours)& \textbf{0.84}  &\textbf{0.15}& 0.07 & \textbf{0.08} & 0.01 &\textbf{18.79} & \textbf{10.65} \\
      \bottomrule
    \end{tabular}
  \end{center}
  \vspace{-12pt}
\end{table*}

\subsubsection{Importance of attention networks}
We use ablations to evaluate the contribution of RH and HH attention networks to the performance of \modelName, as shown in Table~\ref{tab:ablation_results}. 
A visualization of attention weights is in Fig.~\ref{fig:attn_visual} in Appendix.

First, in terms of all metrics in most environments, the model with both RH and HH attention shows the strongest results, followed by the models with only one attention, and finally by the model with no attention. 
For example, in Fig.~\ref{fig:sim_more_crowded}, the ablations in (a), (b), and (c) fail yet our full model in (e) succeeds.
Especially in the 2 most challenging \textit{More crowded} and \textit{More constrained} environments, the existence of either RH or HH attention significantly boosts the average success rate by 0.38 and 0.43 respectively, compared with No attn. 
For example, in Fig.~\ref{fig:sim_more_constrained} (e), the robot detours and goes dangerously close to the crowds in a constrained area, yet HH attention allows it to realize the danger and resume to a less crowded path.  
This result shows that reasoning about both HH and RH spatial relationships is essential for our problem, especially in \textit{more crowded} or \textit{more constrained} environments where the spatial interactions are also dense. 

Second, by comparing RH and HH, we find that HH attention plays a more important role in all environments in terms of success rate and navigation time. 
An extreme example can be observed from \textit{more constrained} environment, where the success rate of RH+HH and HH are the same. 

This is because the number of HH edges is larger than the number of RH edges in most cases. Thus, the robot can observe and is affected by a relatively larger number of HH interactions yet a smaller number of RH interactions. 


\subsubsection{Failure cases}
By visualizing the testing episodes across all environments, we find that \modelName typically collides when (1) a human arrives at its goal and suddenly changes directions to a new goal (Fig.~\ref{fig:failure}(a) in Appendix) or (2) the robot surroundings are extremely crowded and constrained, and almost all free paths are blocked by humans (Fig.~\ref{fig:failure}(b)).
In these cases, due to its speed limit, the robot sometimes cannot switch to an alternative path in time to prevent collisions. 
To remedy the difficulty of RL in long-term decision making, in future work, our method can be combined with motion primitives and path planning algorithms that consider the stochasticity of pedestrian motions~\cite{hu2025composablenav}.

Besides answering the four research questions in Sec. V,
we provide additional insights in Appendix\ref{sec:append_sim}.

%% file: Sections/06-RealExp.tex
\begin{figure*}[ht]

    \centering
    \includegraphics[width=\linewidth]{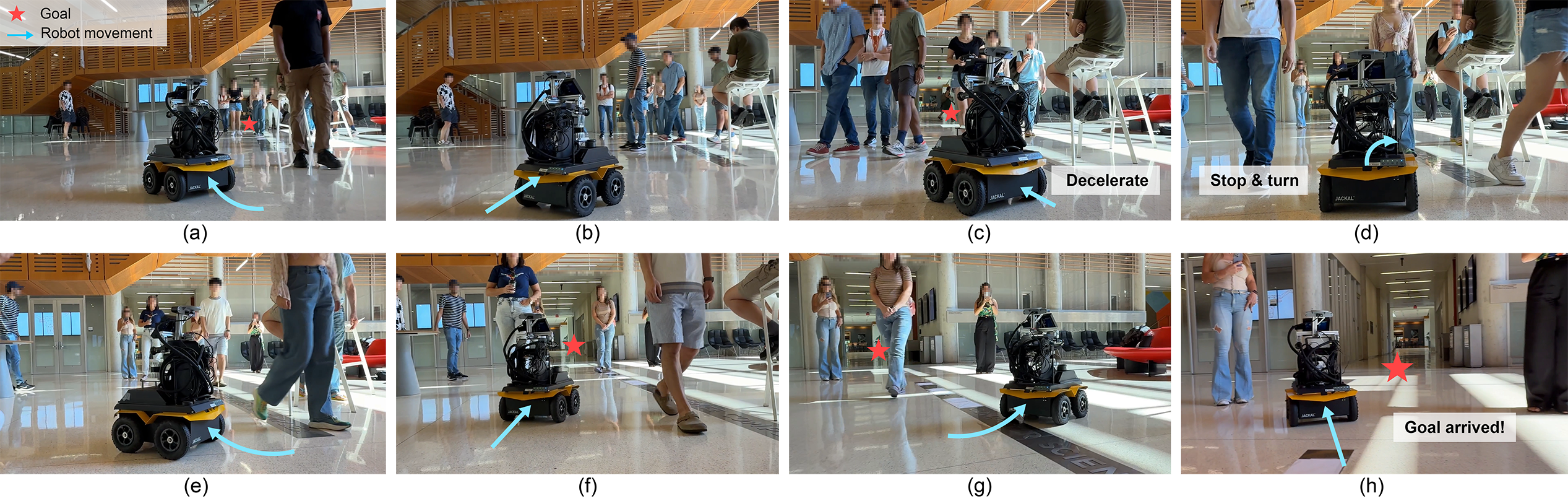}
    
    \vspace{-5pt}
    \caption{\textbf{A deployment scenario of our method in the Atrium environment with 14$\sim$25 humans.} The jackal takes detours to avoid sparse crowds, stops and adjusts its orientation to avoid dense crowds, and arrives at the goal. More qualitative results can be found in the video attachment.}
    \vspace{-5pt}
    \label{fig:real_atrium}
\end{figure*}

\begin{figure*}[ht]

    \centering
    \includegraphics[width=\linewidth]{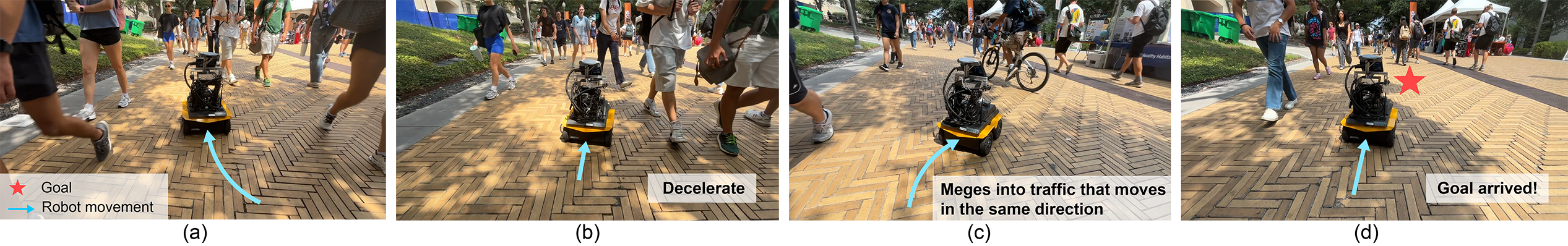}
    
    \vspace{-5pt}
    \caption{\textbf{A deployment scenario of our method in an outdoor environment with 35+ humans.} The jackal takes detours if there is enough space, decelerates if all paths are blocked, and merges into the traffic that moves in the same direction if possible.}
    \vspace{-9pt}
    \label{fig:real_speedway}
\end{figure*}

\section{Real-world Experiments}
\label{sec:real_exp}
In this section, we present challenges, setup, and results in everyday environments in the real-world.

\subsection{Setup}
\label{sec:real_env}
\textbf{Challenges and solutions:} Real-world navigation with pedestrians and static obstacles is challenging because perception errors, unexpected human behaviors, and noisy robot dynamics model cause sim2real gaps.
To deal with noises in human detection, robot localization, and robot dynamics, we inject Gaussian noises into the observed agent positions and the robot dynamics model in the simulator to robustify the learned policy.
To conquer the human behavior gap, in the simulator, we design pedestrian routes based on observations of human flow and the layout of the deployment environment.

We test the policies in four everyday environments \textbf{without any additional training in the real-world}. 
The hallway environment (Fig.~\ref{fig:real_hallway}) has extremely narrow free space and tests the robot's ability to handle static obstacles with low density crowds.
The lounge environment (Fig.~\ref{fig:real_kitchen}) tests the robot's ability to avoid more pedestrians and obstacles with more diverse shapes.
The atrium (Fig.~\ref{fig:real_atrium}) and the outdoor environment (Fig.~\ref{fig:real_speedway}) test the robot's potential of real-world deployment in dense and unrehearsed crowds.
Details about setup can be found in Appendix\ref{sec:append_real}.

In the Hallway and Lounge environments, we conduct controlled experiments where the trajectories of all pedestrians are fixed to ensure a fair comparison.
We compare our method with the following two baselines: (1) ROS navigation stack is a geometric planner which uses the dynamic window approach (DWA)~\cite{fox1997dynamic} as the local planner and $\text{A}^*$ as the global planner;
(2) DRL-VO is a state-of-the-art RL method~\cite{xie2023drlvo}. 
For each method, we run 30 trials in total. Among all trials, the start goal positions of the robot are fixed. 
We measure the success, collision, timeout rates, and navigation time of successful episodes as testing metrics.

To demonstrate the real-world deployment of our method, we conduct in-the-wild experiments in two densely packed public spaces: the Atrium and Outdoor environment in Fig.~\ref{fig:real_atrium} and Fig.~\ref{fig:real_speedway} with non-rehearsed pedestrian flows.

\subsection{Baseline comparison results}

In the highly constrained yet less crowded Hallway environment, ROS navigation stack often needs to spin in place to replan, as shown by the higher navigation time in Table~\ref{tab:real_exp}. Some of the spinning recovery attempts fail and result in timeouts. 
In ROS navigation stack, both global and local planners treat humans as obstacles. As a result, similar to the baselines in Sec.~\ref{sec:result_rep}, the robot has difficulties distinguishing dynamic obstacles that will clear out and static obstacles that will always stay, causing failures in highly constrained spaces. 
For the same reason as simulation experiment in Sec.~\ref{sec:sim_exp}-C-1), the large input representation and network of DRL-VO pose challenges for RL exploration in this highly constrained environment. Insufficient exploration leads to an underfitting problem, where the robot collides with obstacles or goes back and forth without arriving at the goal.
On the contrary, by using different representations of humans and obstacles, our method is able to explore the different strategies to avoid them through trial and error during RL training. 
For example, in Fig.~\ref{fig:real_hallway}(e)-(g), when turning into the narrow doorway, the robot takes a wide turn to yield enough space so that the human can exit the doorway. However, the robot did not exhibit such prediction-aware behaviors when avoiding obstacles.
Consequently, the robot demonstrates high success rates in Table~\ref{tab:real_exp}. Qualitatively, the robot is able to safely pass a human in a narrow corridor (Fig.~\ref{fig:real_hallway} (a)-(d) in Appendix), takes a wide turn to give a human enough room, and enters an extremely narrow doorway (Fig.~\ref{fig:real_hallway} (e)-(h)). 

\begin{table*}[t]
\begin{center}
    \caption{Real-world baseline comparison results}
    \label{tab:real_exp}
    \begin{tabular}{c c l c c c c}
    \toprule
     \textbf{Environment}  &\textbf{ \# of trials}& \textbf{Method} & \textbf{Success}$\uparrow$ & \textbf{Collision}$\downarrow$ & \textbf{Timeout}$\downarrow$ & \textbf{Nav Time}$\downarrow$ \\
     \midrule
     
     \multirow{3}{*}{\shortstack{\textbf{Hallway}\\1-2 humans}}  &\multirow{3}{*}{18}& Navigation Stack   & 0.72& 0.06& 0.22& \textbf{16.71}\\
    && DRL-VO & 0.50 & 0.50 & \textbf{0.00}& 29.33\\
       && \modelName (ours) & \textbf{1.00}& \textbf{0.00}& \textbf{0.00}& 22.36\\
      \midrule
     \multirow{3}{*}{\shortstack{\textbf{Lounge}\\1-6 humans}}  &\multirow{3}{*}{12}& Navigation Stack & \textbf{0.83}& 0.17& 0.00& 32.00\\
     && DRL-VO & 0.67 & 0.33 & 0.00& \textbf{28.91}\\
       && \modelName (ours) & \textbf{0.83}& 0.17& 0.00& 30.71\\
      \bottomrule
    \end{tabular}
\end{center}
\vspace{-15pt}
\end{table*}

In the less constrained yet more crowded Lounge environment, 
DRL-VO has the highest collision rate due to its underfitting problem. In addition, since only DRL-VO requires human velocities as inputs, the noises of human velocity estimation exaggerates its problem.
ROS navigation stack and ours have the same success and collision rate, because the Lounge environment has enough space for the navigation stack to apply a ``stop and wait'' strategy for humans and obstacles. 
However, as a cost, ``stop and wait'' takes a longer average time to arrive at goals as shown in the last column of Table~\ref{tab:real_exp}.
Different from this na\"{\i}ve strategy, \modelName adapts its navigation behaviors based on different types of spatio-temporal interactions. 
For example, in Fig.~\ref{fig:real_kitchen} (a)-(c) in Appendix, since pedestrians walking in opposite directions are crossing each other, the robot first turns left to avoid the lady on its right, and then turns right to avoid the two males on its left. Then, in Fig.~\ref{fig:real_kitchen} (d)-(e) and (f)-(g), the robot chooses actions that deviate from the current and intended paths of pedestrians, walls, and the table to safely arrive at the goal. 
The failure cases of both methods are caused by changes in the arrangement of tables and chairs, while the fake point clouds are generated from a map with outdated furniture positions. This issue can be mitigated by replacing fake point clouds with real-time point clouds from sensors.


\subsection{Deployment results}
The behavior mismatch between ORCA humans in simulation and real pedestrians is exaggerated during deployment. Since real pedestrians are usually risk-averse and conservative when interacting with the robot, all simulated humans react to the robot with a larger safety space. If the simulated humans out of robot's view are about to collide with the robot, they freeze until they are far enough from the robot. 
To conquer perception and dynamics gaps, we strictly duplicate ZED 2i camera's sensor range (120$^{\circ}$ field-of-view (FoV) and humans are only detectable 1$\sim$15 meters from the camera) and use a kinematic bicycle model with measured delays in simulation.

As shown in Fig.~\ref{fig:real_atrium} and Fig.~\ref{fig:real_speedway}, during deployment, our method has learned an adaptive policy that accommodates the robot's limitations. When the crowd is sparse, the robot circles around pedestrians with a large safety distance to keep them in its FoV as long as possible. When the crowd is dense, the robot stops temporarily and keeps adjusting its orientation to find a direction with enough space to move forward, and resumes right after the pedestrians pass the robot. 
In the outdoor environment where pedestrians form a two-way traffic, the robot shows a smart strategy by merging into the traffic that moves in the same direction.
These observations demonstrate that our method has learned an adaptive policy that is robust to sim2real gaps and can navigate safely and efficiently with unpredictable humans.

%% file: Sections/07-Discussion.tex
\section{Discussions, Limitations, and Future Work}
\label{sec:discussion}
In this section, we reflect on the key components of our framework, discuss the limitations of our approach, and propose directions for future research.
\subsection{Sim2real through Real2sim}
\textbf{Lessons learned:}
To overcome sim2real gaps, the design of simulation pipelines needs to be guided by the constraints of hardware and environments in the real-world.
On one hand, we determine the input representation of \modelName based on what could be obtained from sensors and perception modules and how accurate they are. 
We find that intermediate features, such as detected human positions and processed point clouds, reduce sim2real gaps.
On the other hand, we also ensure the consistency of the simulation and real-world, such as the robot dynamics model, whenever possible. 
These design choices allow our policy to generalize to different simulation environments and be deployed in challenging real-world scenarios. 

\textbf{Limitation and future work:}
Although we have minimized the sim2real gaps through real2sim, certain gaps still exist. 
Before sim2real transfer, the policy needs to be finetined in simulations that mimic the target environments (Fig.\ref{fig:sim_env_appendix}). 
To further align the simulation and the real-world, we plan to explore the following directions for future work:  (1) developing a more natural pedestrian model to replace the ORCA humans in the simulator, (2) revising our pipeline to enable failure detections and self-supervised RL fine-tuning in the real-world~\cite{Zhu2020The, chang2023data}, (3) using a small amount of real-world data to automatically optimize the parameters of our simulator to match the real-world environment~\cite{du2021auto,lim2022real2sim2real}.

\subsection{Scene representation}
\textbf{Lessons learned:}
A good scene representation should be tailored to the needs of its downstream task.
Besides the above sim2real considerations, our scene representation is split due the different nature of humans and obstacles for robot collision avoidance. 
The size of humans are small and their shapes are simple. Therefore, to avoid humans, the robot only needs to treat them as moving circles.
In contrast, obstacles have larger and more complicated shapes. The part of the obstacles that faces the robot is more important for collision avoidance. 
Therefore, point clouds are the most intuitive way to represent obstacles.
Our experiments in Sec.~\ref{sec:sim_exp} show that this split scene representation reduces robot collision avoidance with both dynamic and static obstacles.

\textbf{Limitation and future work:}
A side effect of our scene abstraction is the loss of detailed information such as gaits of humans and 3D shapes of obstacles. 
Another side effect is the cascading errors between the perception modules and robot policy, such as inaccurate human detections or robot localization. 
A future direction is the joint optimization of the whole robotic stack from perception to control~\cite{hu2023_uniad,wang2024mosaic}.

\subsection{Structured neural network}
\textbf{Lessons learned:}
Model-based approaches require low-fidelity data yet heavily rely on assumptions. In contrast, end-to-end learning approaches need few assumptions yet require high-fidelity data.
Our structured learning method combines the best of both worlds: It requires low-fidelity data yet relies on minimal assumptions.
By injecting structures into the neural network, we decompose a complex problem into smaller and relatively independent components. Note that our decomposition does not break the gradient flow, which keeps \modelName end-to-end trainable. We propose a principled way for network architecture design, increasing the transparency of these black-boxes. 
Our experiments demonstrate that the structured network outperforms both model-based methods and RL-based methods without structures, which empirically proves the effectiveness of structure learning for interactive tasks with multiple heterogeneous participants.

\textbf{Limitation and future work:}
The structured neural network utilizes the inductive bias behind human behaviors to solve specific problems. However, without re-training, the network cannot generalize to other tasks such as autonomous driving.
In other fields such as computer vision and natural language processing, foundation models that are trained on large-scale diverse data have shown promise in generalization. Inspired by the standardized benchmarks of autonomous driving and robot manipulation~\cite{caesar2021nuplan,openxembodiment2023}, we aim to contribute to the benchmarking efforts of human-centered tasks.

\subsection{Training method}
\textbf{Lessons learned:}
Deep RL is a promising tool to solve robotic problems that are beyond the capabilities of traditional rule-based methods without large-scale real-world datasets.  
It enables the robot to explore the environment and learn meaningful behaviors through trial and error.
Without heuristics or demonstrations, the robot has to collect reward signals to improve its policy. 
Consequently, simulation development with a randomization scheme and a good reward design are indispensable to the performance of our method.  

\textbf{Limitation and future work:}
Our method is subjective to the inherent limitations of RL, such as the need for simulators, low training data efficiency, and difficulties in long-horizon tasks.
In addition, preventing performance degradation of the policy when inevitable distribution shifts happen, especially in real-world, is challenging. 
In future work, combining the RL policy with other planners, imitation learning, or foundation models may alleviate these problems~\cite{raj2024rethinking}.

\subsection{Miscellaneous}
 \textbf{Multi-robot navigation:} This work focuses on single-robot navigation. Modeling robot-robot interactions in multi-robot navigation in crowded and constrained environments is an interesting future direction. To enable this, the extra robots and the interactions among all robots can be treated as extra nodes and spatial edges in the st-graph. Following our method, from the st-graph, robot-robot interactions can be modeled with self-attention or other methods, which serves as another submodule in each robot's policy network.
 
\textbf{Social awareness:} The attention-based interaction framework presents a promising foundation for some social behaviors (e.x. treating humans and obstacles differently). However, \modelName focuses on optimizing task success and efficiency, and thus lacks explicit mechanisms and evaluation metrics for social awareness~\cite{francis2025principles}. Future work could integrate social norm as constraints or auxiliary objectives in RL training, as well as learning from human demonstrations or preference-based feedback. User studies can evaluate social awareness.

%% file: Sections/08-Conclusion.tex
\section{Conclusion}
\label{sec:conclusion}

In this article, we proposed HEIGHT, a novel structured graph network architecture for autonomous robot navigation in dynamic and constrained environments. Our approach takes advantage of the graphical nature and decomposability of the constrained crowd navigation problem, introducing the following two key novelties. First, we split the human and static obstacle representations, which allows the robot to effectively reason about different collision avoidance strategies to deal with both. Second, we propose a heterogeneous st-graph to capture various types of interactions among the robot, humans, and obstacles. The \hstGraph guides the design of the HEIGHT network. Attention mechanism enables the robot to reason about the relative importance of each pairwise interaction, leading to adaptive and agile robot decisions during navigation. 
Our simulation benchmark shows that HEIGHT outperforms model-based and learning-based methods in terms of collision avoidance and navigation time. HEIGHT also demonstrates improved generalization when human and obstacle densities vary. HEIGHT is seamlessly transferred from simulation to real-world navigation scenarios, showcasting its robustness and deployability. 

Our work suggests that reasoning about subtle spatio-temporal interactions is an essential step toward smooth human-robot interaction. Furthermore, our work highlights the significance of uncovering the inherent structure of complex problems and injecting these structures into learning frameworks to solve the problems in a principled manner.

\section*{Acknowledgements}
We thank Zhe Huang for helpful discussions and the code of the A* planner. We thank Chen Tang, Aamir Hasan, and Rutav Shah for feedback on paper drafts. We thank colleagues at University of Illinois Urbana-Champaign and at the University of Texas at Austin who participated in real-world experiments.
This material is based upon work supported by the National Science Foundation (NSF) CAREER-2143435, NSF CAREER-2046955, NSF SLES IIS-2416461, Amazon, and JP Morgan. Any opinions, findings, and conclusions expressed in this material are those of the authors and do not necessarily reflect the views of the sponsors.

%% file: Sections/Appendix.tex
\appendices
\section*{Appendix}
\addcontentsline{toc}{section}{Appendix}

\subsection{Simulation experiments -- Baselines}
\label{sec:append_baseline}
\begin{itemize}
    \item \textbf{Dynamic window approach (DWA)}~\cite{fox1997dynamic} searches for the optimal velocity that brings the robot to the goal with the maximum clearance from any obstacle. DWA is a model-based method that only considers the current state. DWA represents both humans and static obstacles as groups of small circles. We re-mapped the output of DWA to match the turtlebot kinematics and our action space.
    \item \textbf{Optimal Reciprocal Collision Avoidance (ORCA)~\cite{van2011reciprocal}} optimizes for the optimal velocity that brings the robot to the goal by assuming that agents avoid each other under the reciprocal rule. ORCA is a model-based method that only considers the current state. In addition, ORCA represents both humans as circles and static obstacles as line segments. We re-mapped the output of ORCA to match the unicycle kinematics of our robot.
    \item \textbf{\astarCNN}~\cite{perez2021robot} is a hybrid method with sample-based planning and RL. With a map of the environment,  $\textrm{A}^{*}$ is the global planner and generates 6 waypoints in the beginning of an episode. The inputs to the RL policy are a 2D LiDAR point cloud, the robot state, the waypoints, and the robot goal. No human detections are used and human features are included in the point cloud. The point cloud is passed through a CNN, whose output is concatenated with the embedding of other inputs. In addition to the reward function in Eq.~\ref{eq:reward}, the robot receives a heading direction award if it moves towards the goal while being collision-free. 
    \item \textbf{Decentrialized structural RNN (DS-RNN)}~\cite{liu2020decentralized} is an RL-based method that represents static obstacles as groups of small circles. In network architecture, DS-RNN only contains the RH attention and the GRU. 
    \item \textbf{DRL-VO}~\cite{xie2023drlvo} is a hybrid method with a pure pursuit algorithm as the global planner and an RL policy as the local planner. The inputs to the planners are the positions and velocities of humans encoded in an occupancy map, a 2D LiDAR point cloud, and the robot state. The RL network is a CNN without explicit interaction modeling.  
    \item \textbf{Homogeneous graph attention network (\homoBaseline)}~\cite{velickovic2018graph} is RL-based and splits human and obstacle inputs. However, \homoBaseline does not differentiate between 3 types of nodes and 3 types of edges in policy network. Instead, \homoBaseline uses a single self-attention network to weigh humans, the robot, and the obstacle point cloud and feed the weighted sum of all embeddings into a GRU. \homoBaseline is similar to Chen et al.~\cite{chen2019relational} and Liu et al.~\cite{liu2024sample} but the input, output, and training algorithm are kept the same as our method for a fair comparison.
\end{itemize}

\begin{figure}
\centering
\includegraphics[width=\linewidth]{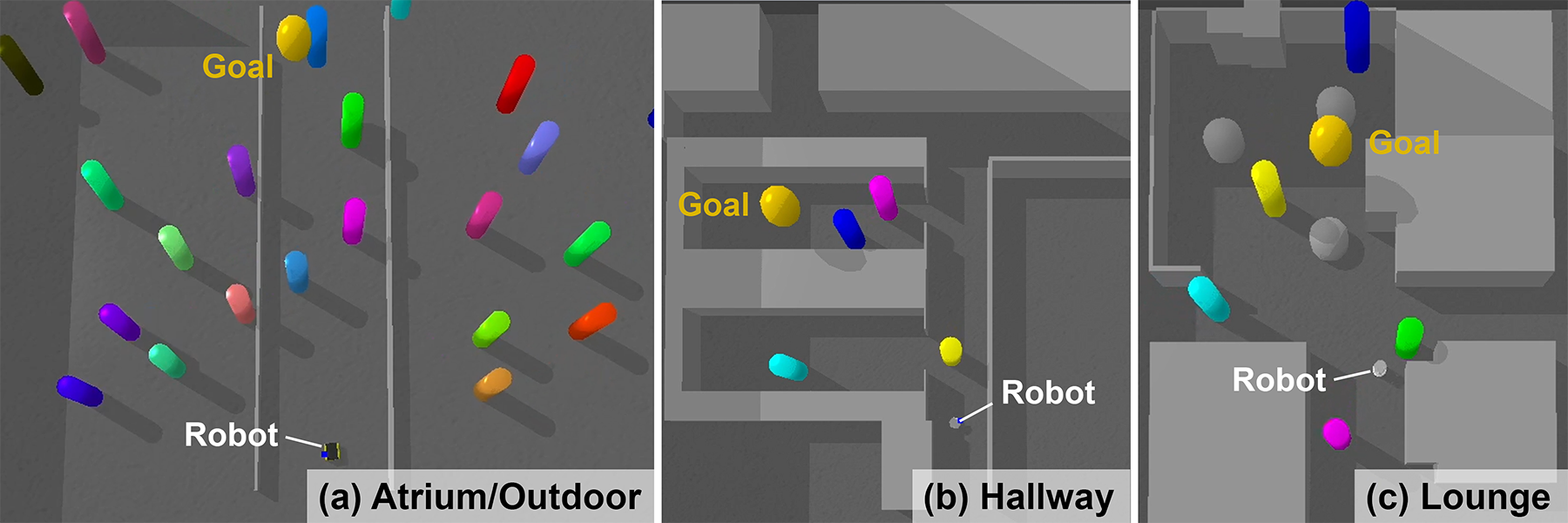}
\caption{Simulation environments for training sim2real policies in Sec.~\ref{sec:real_exp}. 
}
\label{fig:sim_env_appendix}
\vspace{-10pt}
\end{figure}

 \begin{figure*}[ht]

    \centering
    \includegraphics[width=\linewidth]{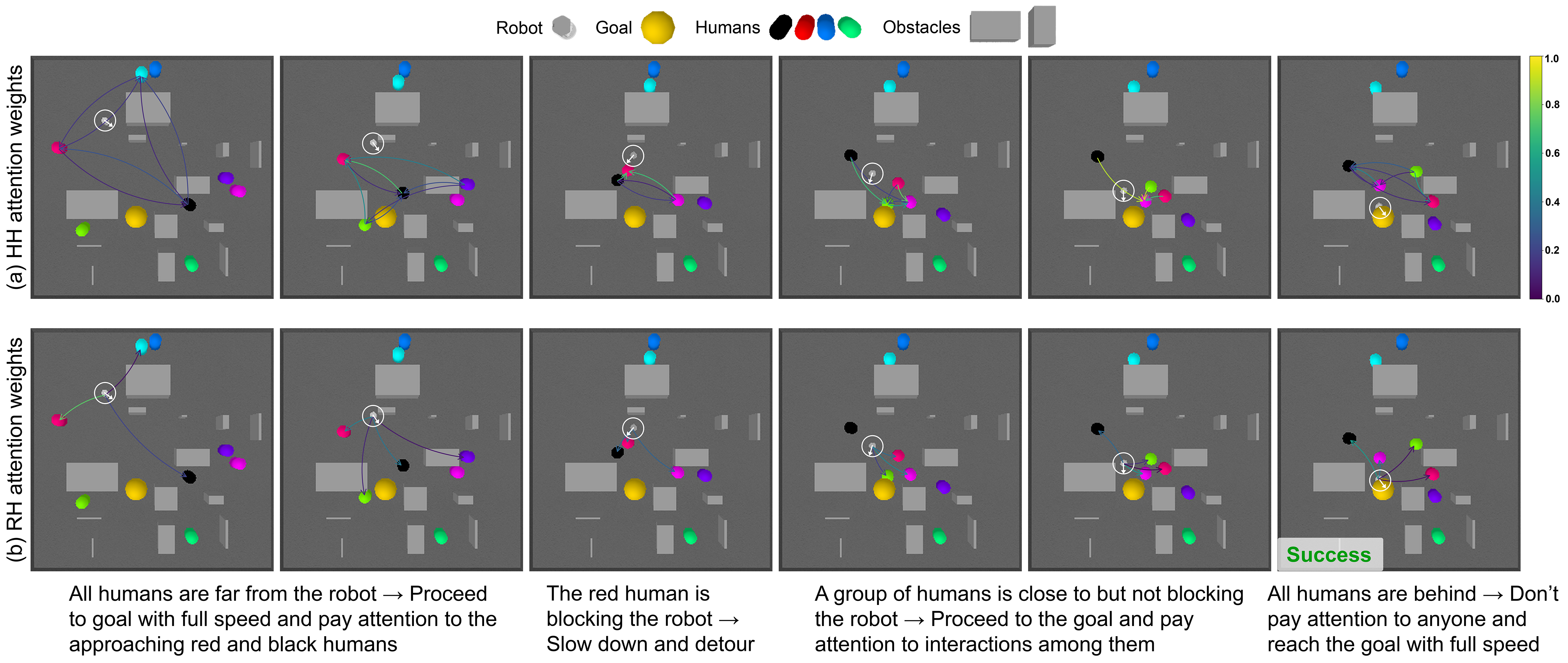}
    
    
    \caption{\textbf{Visualizations of HH and RH attention weights of HEIGHT.} The color of lines indicate the weight values (see the colorbar on the right). For simplicity, we only include the line if the weight value is greater than 0.1. We observe that the interactions that likely to affect robot navigation in the near future have high attention values. Examples include humans that are approaching the robot and humans that block the robot's path to the goal.}
    \vspace{-5pt}
    \label{fig:attn_visual}
\end{figure*}

 \begin{figure*}[ht]

    \centering
    \includegraphics[width=\linewidth]{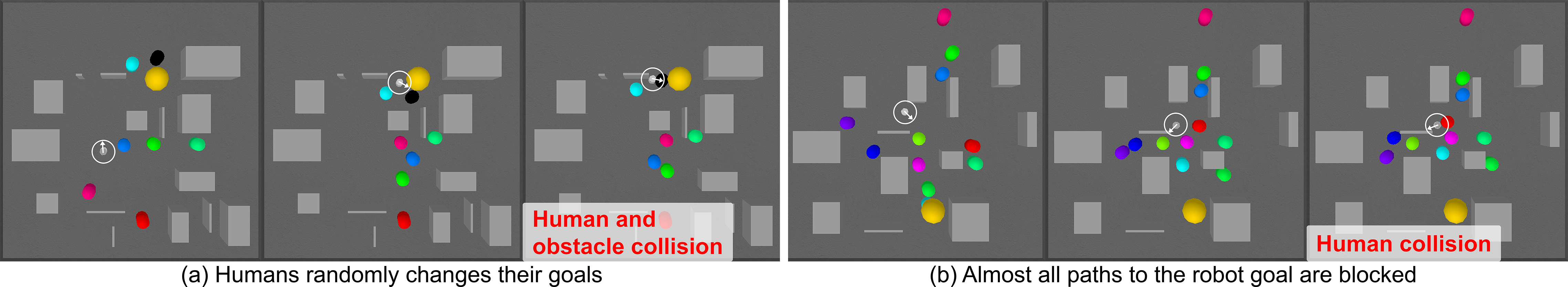}
    
    
    \caption{\textbf{Failure cases of our method.} (a) The black human was going downwards for a while, but suddenly changes its direction toward the robot. (b) All efficient paths toward the goal are blocked or will soon be blocked by human crowds.}
    \vspace{-10pt}
    \label{fig:failure}
\end{figure*}

\begin{figure*}[ht]

    \centering
    \includegraphics[width=\linewidth]{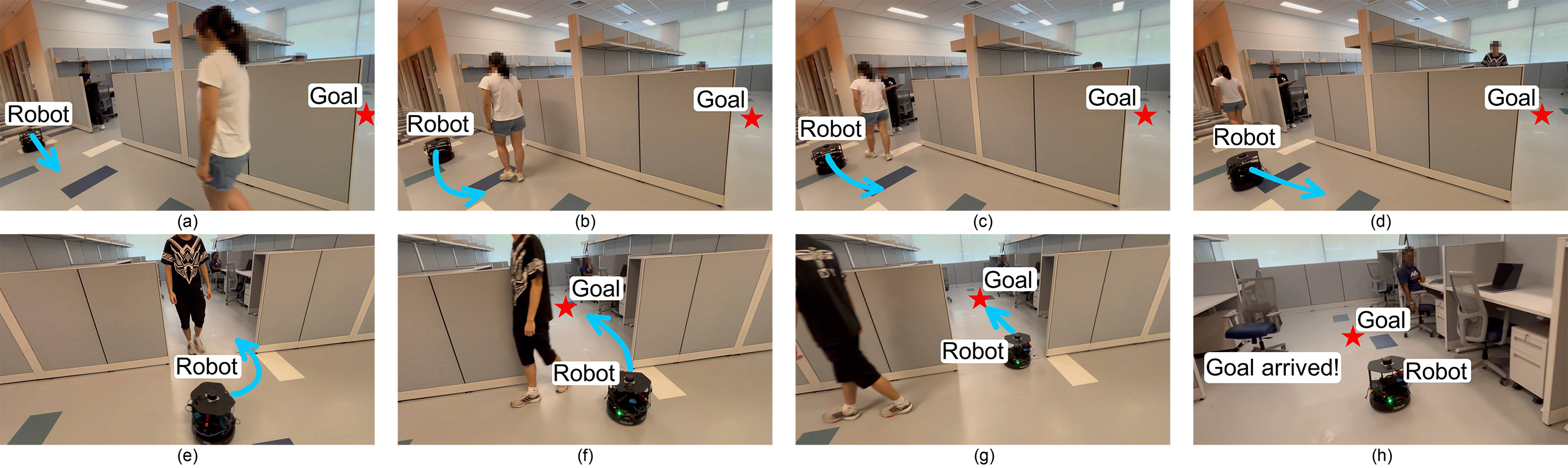}
    
    \caption{\textbf{A testing episode of our method in the real Hallway environment.} The blue arrow denotes the robot path that results from its actions. The red star denotes the goal position. The turtlebot avoids two pedestrians, one after another in a narrow corridor, enters a narrow doorway, and arrives at the goal. }
    \vspace{-5pt}
    \label{fig:real_hallway}
\end{figure*}

\begin{figure*}[ht]

    \centering
    \includegraphics[width=\linewidth]{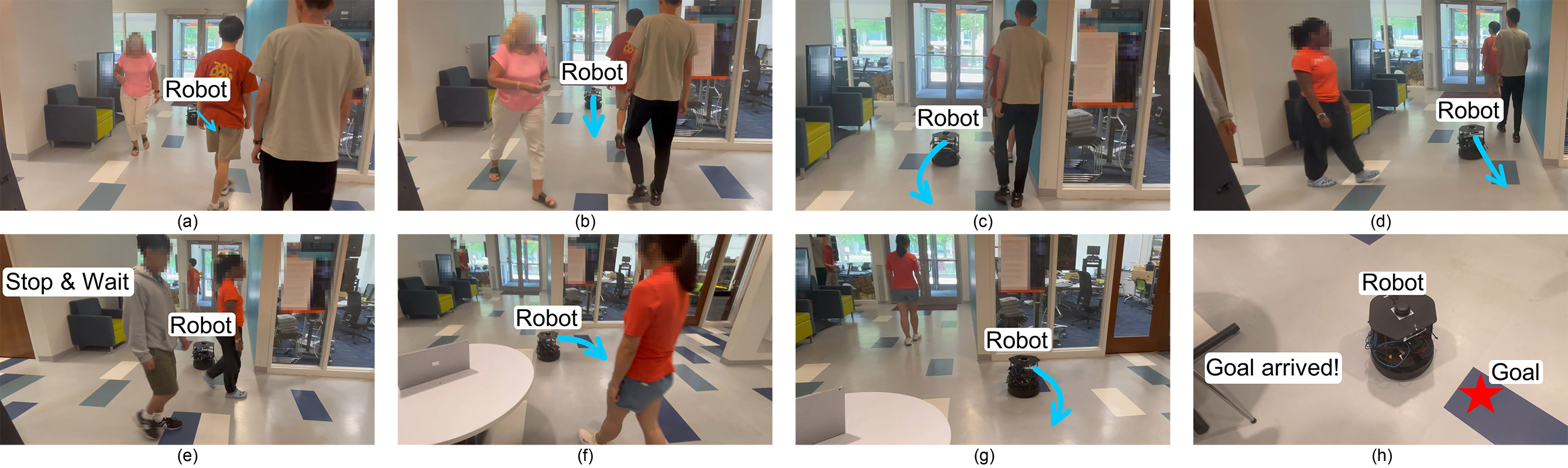}
    
    \caption{\textbf{A testing episode of our method in the real Lounge environment.} The turtlebot avoids multiple groups of people who pass each other in different heading directions, avoids the walls and furnitures, and arrives at the goal. 
    }
    \vspace{-10pt}
    \label{fig:real_kitchen}
\end{figure*}

\subsection{Simulation experiment setup}
\label{sec:append_resource}
\subsubsection{Training details}
We train all RL methods, including all baseines and ablations except DWA, for $2\times 10^8$ timesteps with a learning rate $ 5\times 10^{-5}$. The learning rate decays at a linear rate with respect to training timesteps.  
To accelerate and stabilize training, we run $28$ parallel environments to collect the robot's experiences. At each policy update, 30 steps of 6 episodes are used. 

\subsubsection{Resource usage}
In simulation experiments, we train and test all methods in a commercial desktop computer with a Ubuntu 20.04 OS, an Intel Core i9-13900K processor, 32 GB RAM, and a NVIDIA RTX 4090 GPU. 
We report the training and testing time in simulation with the above computer in Table~\ref{tab:runtime}. For training time, all RL methods except DRL-VO are similarly fast, while DRL-VO is the slowest due to its high-dimensional occupancy map input and the larger 2D CNN network. 
For inference time, \modelName is the slowest because the two attention networks for spatial reasoning and the GRU for temporal reasoning increases computation cost. Thus, we observe a tradeoff between performance and time efficiency. In future work, the inference latency can be reduced by (1) combining the state of multiple humans in the same group and (2) deploying transformer acceleration techniques such as key-value caching or sparse attention.

    \begin{table}[t]
  \centering
  \caption{Comparison of training and inference time in the training environment. All RL methods are trained for $2\times 10^8$ steps. The inference time is averaged over 500 testing steps.}
  \label{tab:runtime}
  \begin{tabular}{lcc}
    \toprule
    \textbf{Method} & \textbf{Training (hours)} & \textbf{Inference (ms/step)} \\
    \midrule
    DWA~\cite{fox1997dynamic} & N/A & 1.5 \\
    ORCA~\cite{van2011reciprocal} & N/A & 0.3 \\
    \astarCNN~\cite{perez2021robot} & 47.2 & 2.0 \\
    DS-RNN~\cite{liu2020decentralized} & 49.0 & 1.0 \\
    DRL-VO~\cite{xie2023drlvo} & 111.9 & 2.3 \\
    \homoBaseline~\cite{velickovic2018graph} & 47.7 & 2.2 \\
    \modelName (ours) & 49.7 & 2.4 \\
    \bottomrule
  \end{tabular}
  \vspace{-15pt}
\end{table}

\subsection{Simulation experiments -- additional insights}
\label{sec:append_sim}
\textbf{Effectiveness of RL planning:}
In Table~\ref{tab:baseline_results}, DWA and ORCA are model-based approaches that determines robot actions based on only the current state, which leads to the worst performances. Thus, reactive policies are not sufficient to solve our problem, justifying the necessity of long-sighted planning.
As a step forward, \astarCNN combines planning and RL, yet the planner relies on occupancy maps without humans, reducing the optimality of planned waypoints and negatively affecting overall navigation. On the other hand, RL learns to maximize the expected long-term returns based on both the current and historical states of all components in the scene. Consequently, the remaining 3 RL methods, especially \homoBaseline and \modelName, outperform \astarCNN in most metrics and in most environments. 
Thus, to ensure the best navigation performance, it is important to optimize the entire planning system based on the task instead of optimizing only a part of it.

\textbf{Difficulty of scenarios:}
From Table~\ref{tab:baseline_results}
 and Table~\ref{tab:ablation_results}, we observe that, besides the overfitted DS-RNN, all methods show the same trend of performance change in the 5 environments: \textit{Less crowded} $>$ \textit{Less constrained} $>$ \textit{More constrained} $>$ \textit{More crowded}.  Obviously, adding more humans or obstacles increases task difficulty. But interestingly, the change in the number of humans has a larger effect on the task difficulty than the change in the number of obstacles. This phenomenon shows that avoiding collisions with humans is more difficult than avoiding obstacles in nature.

\subsection{Real-world experiment setup}
\label{sec:append_real}
We train the sim2real policies in the simulators of a hallway (Fig.~\ref{fig:sim_env_appendix}(b) in Appendix), a lounge (Fig.~\ref{fig:sim_env_appendix}(c)), and an atrium (Fig.~\ref{fig:sim_env_appendix}(a)) for $2\times 10^8$ timesteps with a decaying learning rate $ 5\times 10^{-5}$. 
The distance between the starting and the goal position of the robot ranges from 6m to 11m. 
The pedestrians were told to react naturally to the robot based on their own preferences. In some testing episodes, other pedestrians who were unaware of our experiment also engaged with the robot.

We use the following two sets of robot hardware:
\begin{itemize}
    \item In Hallway and Lounge environments, 
we use a TurtleBot 2i equipped with an RPLIDAR-A3 laser scanner, an Intel RealSense tracking camera T265, a Nvidia Jetson Xaiver board, and a remote host computer.
We first use the ROS \texttt{gmapping} package to create a map of the environment. Then, we process the map to combine small obstacles and eliminate noises. 
To detect human positions from LiDAR point clouds, we use an off-the-shelf people detector, DR-SPAAM~\cite{Jia2020DRSPAAM}. 
Since DR-SPAAM tends to treat obstacles as false positive humans, we mitigate this issue by removing the point clouds that belong to pre-mapped static obstacles from the LiDAR readings, then feed the remaining point clouds that only contain humans to DR-SPAAM.
We use the T265 camera to obtain the pose of the robot $(p_x, p_y, \theta)$ and the robot wheel encoder to obtain its velocity $(u_x, u_y)$.
The host computer runs the robot policies and perception modules, while the TurtleBot only receives and sends ROS messages. The host computer and TurtleBot communicate through ROS.

\item In Atrium and Outdoor environment, 
we use a Clearpath Jackal robot equipped with an ZED 2i camera and an Ouster LiDAR. 
The system is built using ROS.
The software stack is similar to the TurtleBot except the following differences:
We use ZED 2i to detect the positions of humans and the pose of the robot. The pre-built map and robot localization are obtained from ENML algorithm~\cite{enml}. 
All computations are executed entirely onboard, utilizing an Intel i7-9700TE CPU and an NVIDIA RTX A2000 GPU. 

\end{itemize}

